\documentclass[sigplan,screen]{acmart} %

\usepackage{balance}
\usepackage{multicol}
\usepackage{microtype}
\usepackage{graphicx}
\usepackage{subfigure}
\usepackage{xspace}
\usepackage{xcolor}
\usepackage{booktabs}
\usepackage{pifont}
\usepackage{fvextra}
\usepackage{amsmath}
\usepackage{enumitem}
\usepackage{algorithm}
\usepackage[noend]{algpseudocode}
\usepackage{adjustbox}
\usepackage[font=small,labelfont={small,bf}]{caption}
\usepackage[font=small,labelfont={small,bf}]{subcaption}
\usepackage{bm}
\usepackage{relsize}
\usepackage{tikz}

\author{Yilong Zhao}
\authornote{Both authors contributed equally to this work.}
\affiliation{%
  \institution{University of California, Berkeley}
  \city{Berkeley}
  \state{CA}
  \country{USA}
}
\email{yilongzhao@berkeley.edu}

\author{Shuo Yang}
\authornotemark[1]
\affiliation{%
  \institution{University of California, Berkeley}
  \city{Berkeley}
  \state{CA}
  \country{USA}
}
\email{andy_yang@berkeley.edu}

\author{Kan Zhu}
\affiliation{%
  \institution{University of Washington}
  \city{Seattle}
  \state{WA}
  \country{USA}
}
\email{kanzhu@cs.washington.edu}

\author{Lianmin Zheng}
\affiliation{%
  \institution{University of California, Berkeley}
  \city{Berkeley}
  \state{CA}
  \country{USA}
}
\email{lianminzheng@gmail.com}

\author{Baris Kasikci}
\affiliation{%
  \institution{University of Washington}
  \city{Seattle}
  \state{WA}
  \country{USA}
}
\email{baris@cs.washington.edu}

\author{Yifan Qiao}
\affiliation{%
  \institution{University of California, Berkeley}
  \city{Berkeley}
  \state{CA}
  \country{USA}
}
\email{yifanqiao@berkeley.edu}

\author{Yang Zhou}
\affiliation{%
  \institution{University of California, Davis}
  \city{Sacramento}
  \state{CA}
  \country{USA}
}
\email{yangzhou.rpc@gmail.com}

\author{Jiarong Xing}
\affiliation{%
  \institution{Rice University}
  \city{Houston}
  \state{TX}
  \country{USA}
}
\email{jxing@rice.edu}

\author{Ion Stoica}
\affiliation{%
  \institution{University of California, Berkeley}
  \city{Berkeley}
  \state{CA}
  \country{USA}
}
\email{istoica@berkeley.edu}

\newif\ifdraft
\drafttrue %

\newif\ifdiff
\difffalse

\newcommand{\new}[1]{\ifdiff{\textcolor{blue}{{#1}}}\else#1\fi}

\renewcommand{\emph}[1]{\textit{#1}}

\newcommand{\sys}{BlendServe\xspace}

\newcommand{\density}{compute density}

\newcommand{\lm}{Llama}
\newcommand{\sota}{state-of-the-art}
\newcommand{\offinf}{offline batch inference}
\newcommand{\Offinf}{Offline batch inference}

\newcommand{\inlen}{input length}
\newcommand{\outlen}{output length}
\newcommand{\Outlen}{Output length}

\newcommand{\psr}{prefix sharing ratio}
\newcommand{\Psr}{Prefix sharing ratio}

\newcommand{\avgimprovethansota}{$20.84$\%}
\newcommand{\avgtoopt}{$86.55$\%}
\newcommand{\upimprovethanvllm}{$1.44\times$}

\newcommand{\refsec}[1]{\S~\ref{#1}}

\newcommand{\kv}{KV-cache\xspace}

\newcommand{\MyPara}[1]{\vspace{.2em}\noindent\textbf{#1}~}
\newcommand*\circled[1]{\tikz[baseline=(char.base)]{
            \node[shape=circle,draw,inner sep=0.1pt] (char) {#1};}}

\copyrightyear{2026}
\acmYear{2026}
\setcopyright{cc}
\setcctype{by}

\acmConference[ASPLOS '26] {Proceedings of the 31st ACM International Conference on Architectural Support for Programming Languages and Operating Systems, Volume 2}{March 22--26, 2026}{Pittsburgh, PA, USA.}
\acmBooktitle{Proceedings of the 31st ACM International Conference on Architectural Support for Programming Languages and Operating Systems, Volume 2 (ASPLOS '26), March 22--26, 2026, Pittsburgh, PA, USA}

\acmISBN{979-8-4007-2359-9/2026/03}
\acmDOI{10.1145/3779212.3790133}   %



\renewcommand{\shortauthors}{Yilong Zhao et al.}

\begin{document}

\ifdiff
\thispagestyle{empty}

\begingroup
\centering
\vspace{-1em}
{\Huge\bfseries Summary of Changes \par}
\vspace{1em}
\endgroup

\begin{multicols}{2}

We thank the reviewers for their insightful comments and have incorporated all feedback into the revised paper, which further strengthens and clarifies the work. 
We begin by summarizing the general \textit{revision comments}, followed by responses to feedback from \textit{individual reviewers}, with corresponding references to the paper.

\section*{General Revision Comments}

\MyPara{Expand evaluation scope.}
In addition to the original evaluation on \lm{}-3-8B with $1\times$A100, we provide results on 5 additional models including \textit{Qwen-2.5-7B, Qwen-2.5-72B, DeepSeek-67B, \lm{}-3-70B, and \lm{}-2-7B} (\refsec{sec-eval-subsec-e2e} and \refsec{sec:eval:distributed}). 
For the large-scale distributed setup, we provide evaluation ranging from $2$ to $8$ A100 GPUs, including both tensor parallelism (Figure~\ref{fig:eval-e2e-tps} (b)) and data parallelism (Table~\ref{tab:eval-dp}).

\MyPara{Clarify novelty over prior work.}
We include explicit discussion on DistServe~\cite{zhong2024distservedisaggregatingprefilldecoding} and Orion~\cite{orion} in~\refsec{sec:bg-opts}. We also provide empirical evidence of DistServe's inefficiency on offline setups with comprehensive evaluations in~\refsec{sec-eval-subsec-e2e}. 
We include DistServe (OSDI'24) as a new baseline in addition to vLLM (SOSP'24), SGLang (NIPS'25), and NanoFlow (OSDI'25).

\MyPara{Improve methodology clarity.}
We revise and polish the definitions and derivations as highlighted in~\refsec{sec-analysis}. 
We provide a detailed explanation of \textit{convergence criteria} and \textit{stopping conditions} of the proposed algorithm in~\refsec{sec:design:robustness-analysis}. We also provide a pseudocode of the algorithm to further clarify in~\refsec{sec:appendix-pseudo-algorithm}.

\section*{Individual Review Comments}
\subsection*{Review A}
\MyPara{Evaluation scope.}
We provide results on $5$ more models in~\refsec{sec-eval-subsec-e2e}, and include distributed inference evaluations ranging from $2$ to $8$ A100 GPUs (\refsec{sec-eval-subsec-e2e} and \refsec{sec:eval:distributed}).

\MyPara{Clarification on memory partition.}
We incorporate more discussion on the memory partition algorithm in~\refsec{sec:design:dual-scanner}.

\MyPara{Compatibility to MLA.}
We include analysis and discussions on \sys{}'s compatibility with various attention variants (e.g., GQA, MLA, and GLA) in~\refsec{sec:discussion}.

\subsection*{Review B}
\MyPara{Clarification on performance reasoning.}
We add detailed contexts and rationale behind each derivation as highlighted in~\refsec{sec-analysis}. We explicitly explain each constant involved.

\MyPara{Scheduling granularity.}
We emphasize that all baselines incorporated \textit{continuous batching}, which performs scheduling at \textit{request-level} granularity (\refsec{sec:eval-setup}) rather than \textit{batch-level}. Thus, \sys{} benefits from its scheduling of request ordering rather than schedule granularity.

\MyPara{Detailed ablation.}
We include a performance comparison with NanoFlow-Balance (Figure~\ref{fig:eval-resource}) as suggested by reviewer.

\subsection*{Review C}
\MyPara{Comparison to prior works.}
We include explicit discussion on DistServe~\cite{zhong2024distservedisaggregatingprefilldecoding} and Orion~\cite{orion} in~\refsec{sec:bg-opts}. We also provide empirical evidence of DistServe's inefficiency on offline setups with comprehensive evaluations in~\refsec{sec-eval-subsec-e2e}. 
We include DistServe (OSDI'24) as a new baseline in addition to vLLM (SOSP'24), SGLang (NIPS'25), and NanoFlow (OSDI'25).

\MyPara{Clarification of novelty.}
We emphasize that the key insight of \sys{} is to find a best request ordering that maximizes both resource overlapping and prefix sharing, rather than prefill/decode collocation.

\MyPara{Citation for \outlen{} prediction.}
We include several concurrent works related to \outlen{} prediction as suggested by the reviewer in~\refsec{sec:design:data-structure}.

\MyPara{Evaluation scope.}
We provide results on $5$ more models in~\refsec{sec-eval-subsec-e2e}, and include distributed inference evaluations ranging from $2$ to $8$ A100 GPUs (\refsec{sec-eval-subsec-e2e} and \refsec{sec:eval:distributed}).

\MyPara{Design details.}
We provide a detailed explanation of \textit{convergence criteria} and \textit{stopping conditions} of the proposed algorithm in~\refsec{sec:design:robustness-analysis}.

\subsection*{Review D}
\MyPara{Evaluation scope.}
We provide results on $5$ more models in~\refsec{sec-eval-subsec-e2e}, and include distributed inference evaluations ranging from $2$ to $8$ A100 GPUs, including both tensor and data parallelism (\refsec{sec-eval-subsec-e2e} and \refsec{sec:eval:distributed}).

\MyPara{Compatibility with different parallelisms.}
We add analysis and discussion on \sys{}'s compatibility on distributed parallelisms including DP, TP, PP, SP, and CP (\refsec{sec:discussion}).

\subsection*{Review E}
\MyPara{End-to-end latency constraints.}
We add a discussion on the proposed algorithm's impact on end-to-end latency in~\refsec{sec:discussion}.

\MyPara{Significance of prefix sharing.}
We emphasize prefix sharing as a widely-used \textit{lossless} optimization that works for both compute- and memory-intensive workloads detailed in~\refsec{sec:bg-opts}.

\MyPara{Clarification of formulation $T_o$.}
We include detailed contexts and clarification of $T_o$ as highlighted in~\refsec{sec:formulate-optimal-throughput}.

\end{multicols}
\newpage

\fi

\title[Optimizing Offline LLM Inference with Resource-Aware Batching]{\sys{}: Optimizing Offline Inference with Resource-Aware Batching}

\begin{abstract}

\Offinf{} is gaining popularity as a cost-effective solution for latency-insensitive tasks, such as model evaluation and data curation.
As the latency objective is highly relaxed, maximizing throughput becomes the primary goal in offline inference.
Previous studies focused solely on optimizing throughput within a batch. However, the diverse resource demands (compute-intensive vs. memory-intensive) across a wide range of applications make these approaches less effective, as imbalanced resource demands between batches restrict optimization opportunities.

Our insight for achieving optimal throughput is to reorder requests into batches that mix compute- and memory-intensive workloads to maximize resource overlap. However, such a request schedule can conflict with the schedule that maximizes prefix sharing, a widely-used performance optimization, causing suboptimal inference throughput. 
In this paper, we first build a performance model to analyze request resource demands. Based on it, we design \sys{}, which harmonizes both resource overlapping and prefix sharing to maximize throughput.
\sys organizes all requests using a resource-aware prefix tree and proposes a dual scanning algorithm to obtain the request schedule.
Our evaluation on various models and workloads shows that \sys can achieve up to 90\% of the optimal throughput.

\end{abstract}

\maketitle %

\section{Introduction}
\label{sec:introduction}

\Offinf{} is becoming increasingly popular as a cost-effective solution for Large Language Model (LLM) inference. It processes requests in batches and returns responses within an extended time window, e.g., 24-hour response window offered by OpenAI’s batch APIs~\cite{openai-batch-api}. 
The relaxed latency objective significantly reduces service costs---for example, OpenAI’s Batch API offers inference at half the cost of its online counterpart. 
This cost advantage has made \offinf{} an attractive choice for a wide range of latency-insensitive applications, including model evaluation~\cite{hendrycks2021measuringmassivemultitasklanguage}, data curation~\cite{huggingfaceCosmopediaCreate}, document summarization~\cite{bai2024longbenchbilingualmultitaskbenchmark}, and predictive analytics~\cite{liu2024optimizingllmqueriesrelational}.
Almost all major inference providers offer \offinf{} services today~\cite{anyscaleOfflineBatch,anthropic-batch-api,AWS-offline-inference,Databricks-offline-inference}.

As the latency objective is highly relaxed, \offinf{} providers prioritize optimizing generation throughput, i.e., tokens per second,
which requires maintaining high concurrent utilization of both compute and memory resources.
In transformer-based LLM inference, there are two phases: prefill, which mainly processes input tokens, and decode, which generates output tokens. 
Both phases use the same model weights and operations, but the prefill phase processes tokens in parallel, making it more compute-intensive, while the decode phase generates tokens sequentially, making it more memory-intensive. 
Prior studies have exploited this distinction to improve inference throughput in the context of \textit{online inference}. 
Sarathi-Serve~\cite{agrawal2024tamingthroughputlatencytradeoffllm} proposes \textit{chunked prefill}, which splits large prefill phases into smaller chunks and schedules them alongside decode phases across iterations, improving arithmetic intensity per iteration for higher throughput.
\new{
Orion~\cite{orion} improves utilization with \textit{operator-level scheduling}, which collocates compute- and memory-intensive operators. NanoFlow~\cite{zhu2024nanoflowoptimallargelanguage} further advances this by partitioning a large request batch into nano batches for finer-grained overlapping, achieving \sota{} throughput.
}

\begin{figure}[!t]
    \centering
    \includegraphics[width=0.95\linewidth]{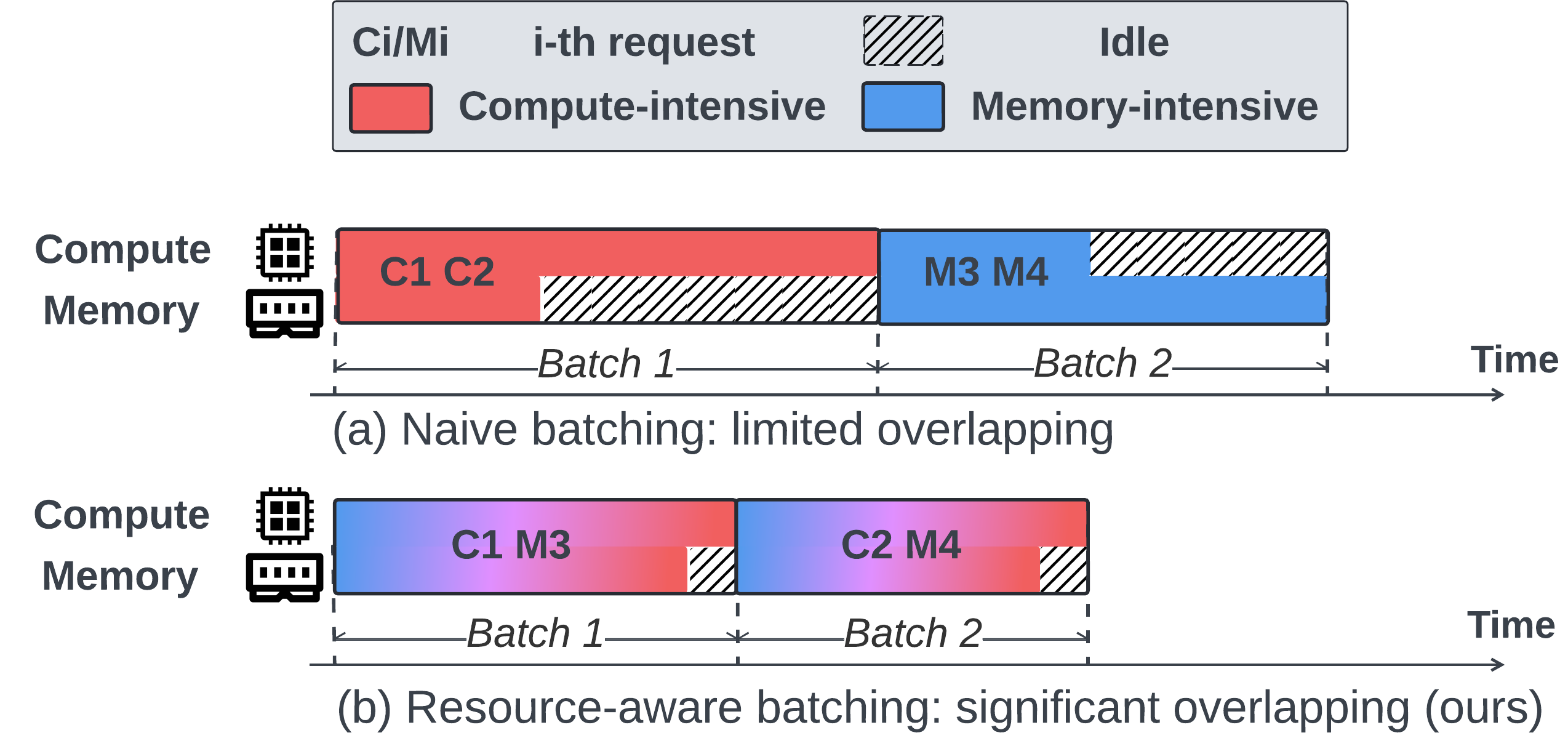}
    \caption{Two ways of batching compute- and memory-intensive requests in offline inference. 
    (a) Naively batching requests in order leads to limited compute-memory overlapping. 
    (b) Resource-aware batching (ours) blends compute- and memory-intensive requests and achieves significant overlapping.
    }
    \label{fig:intro-overview}
\end{figure}

However, these online inference optimizations are far from achieving optimal throughput for \textit{offline scenarios}. 
This is because they only focus on optimizing execution within a request batch but overlook the opportunities across batches, 
which becomes increasingly important as request diversity grows rapidly.
Specifically, advancements in model capabilities have expanded their applications across a wide range of domains, such as chatbots~\cite{chatgpt}, math~\cite{ye2025limoreasoning}, and coding~\cite{githubGitHubCopilot}. Besides, the rise of multi-modal models\cite{wang2024emu3nexttokenpredictionneed, wu2024vilauunifiedfoundationmodel, wang2024miofoundationmodelmultimodal, wu2024janusdecouplingvisualencoding} has further extended their reach to image and video understanding and generation.
Such application diversity leads to numerous requests with \textit{diverse resource demands}. For example, document summarization has long input sequences but short output tokens, which consumes more compute, whereas video generation produces significantly more output tokens, which need more memory bandwidth. 
If a batch is dominated by a single request type (e.g., all compute-intensive), opportunities for overlapping compute and memory-bandwidth usage will be limited, as shown in Figure~\ref{fig:intro-overview}(a).

\MyPara{Insight.}
Our key insight is to carefully construct batches in a resource-aware manner. 
Specifically, by combining (or blending) compute- and memory-intensive requests with a certain ratio to form a batch, we can maximize opportunities for concurrent execution of compute- and memory-intensive operations, enhancing hardware utilization and effectively improving throughput. 
We illustrate this idea in Figure~\ref{fig:intro-overview}(b).

\MyPara{Key challenge.}
However, considering compute-memory overlapping in isolation might not provide optimal throughput, as it usually conflicts with another widely used technique to improve throughput---prefix sharing~\cite{zheng2024sglangefficientexecutionstructured,lin2024parrotefficientservingllmbased,juravsky2024hydragenhighthroughputllminference}.
Prefix sharing group requests with shared prefixes, which allows the shared portion to be computed only once, avoiding redundant computation and \kv{} storage. 
Studies have shown that when optimally utilized---by processing requests in an optimal order---prefix sharing can increase throughput by $6.4\times$ on certain workloads~\cite{zheng2024sglangefficientexecutionstructured}.
However, a request order that achieves high prefix sharing does not necessarily yield high compute-memory overlapping, and vice versa. 
For example, document summarization requests are compute-intensive, but they usually only share the same prefix with other summarization requests, instead of memory-intensive video generation requests; a request order optimizing for prefix sharing would prevent compute-memory overlapping. 
Therefore, we must consider both factors together for maximizing throughput.

\MyPara{\sys{}.}
In this work, we design \sys{}, the first serving system that is specifically optimized for \offinf{} by leveraging both (a) blending compute-intensive and memory-intensive requests, on one hand, and (b) prefix sharing, on the other hand.
We first conduct a deep performance analysis and develop a theoretical model to characterize requests with diverse resource demands.
Based on the model, \sys{} constructs a resource-aware prefix tree, where each node encodes the \density{} of all requests within its subtree. It then sorts the tree nodes based on their density values, placing compute-intensive nodes on the left and memory-intensive nodes on the right. The sorted tree preserves the structure of the prefix tree, so it inherits the benefit of prefix sharing.
To determine the best request order for batching, \sys{} employs a dual scanner algorithm, which scans the tree leaves from left and right simultaneously, effectively batching compute-intensive requests with memory-intensive requests to maximize compute-memory overlapping.
Finally, \sys{} extends the design to data parallelism and tensor parallelism to support large-scale deployment with larger models and clusters~\cite{shoeybi2020megatronlmtrainingmultibillionparameter}.

We prototyped \sys{} based on NanoFlow~\cite{zhu2024nanoflowoptimallargelanguage}, which has integrated chunked prefill~\cite{agrawal2024tamingthroughputlatencytradeoffllm}, and extended it with our resource-aware prefix tree and dual scanner algorithm for optimized batch formulation.
We evaluated \sys{} on a range of models including \lm{}-3-8B, \lm{}-3-70B~\cite{dubey2024llama3herdmodels}, and Qwen-2.5-7B~\cite{bai2023qwentechnicalreport}, and datasets featuring different performance characteristics, including chatbots~\cite{zhao2024wildchat1mchatgptinteraction}, benchmark~\cite{hendrycks2021measuringmassivemultitasklanguage}, API service~\cite{wang2024burstgpt} and vision workloads~\cite{nan2024openvid1mlargescalehighqualitydataset}.
We compared \sys{} against commonly-used systems including vLLM~\cite{vllm}, SGLang~\cite{zheng2024sglangefficientexecutionstructured}, and NanoFlow~\cite{zhu2024nanoflowoptimallargelanguage}. 
Compared to the industry-standard vLLM and SGLang, \sys{} achieves up to \upimprovethanvllm{} throughput speedup. It also delivers an average \avgimprovethansota{} higher throughput than NanoFlow, the current state-of-the-art throughput-oriented inference system.
More importantly, our analysis shows that \sys{} reaches an average \avgtoopt{} (up to 90\%) of the achievable optimal throughput, demonstrating its effectiveness.

In summary, our main contributions include:
\begin{itemize}[itemsep=1pt, topsep=1pt,leftmargin=*]
    \item We conducted a detailed analysis of offline serving workloads and built a performance model to analyze their compute and memory resource demands.
    \item We designed a resource-aware prefix tree for request management that encodes resource demands while preserving prefix structures.
    \item We proposed a request batching algorithm that optimizes throughput by maximizing compute-memory overlapping while preserving high prefix sharing.
    \item We built a prototype and evaluated it comprehensively, demonstrating that it achieves an average \avgtoopt{} (up to 90\%) of the optimal throughput.
\end{itemize}

\section{Background}
\label{sec:background}

\subsection{Transformer-based large model inference}
\label{sec:bg:basic-workflow}

\MyPara{Transformer-based LLM.} 
The core of transformer is its self-attention mechanism, which enables a model to capture the dependencies between all tokens in a sequence.
This is achieved via query (Q), key (K), and value (V) transformations, where each token's embedding is projected into Q, K, and V tensors. The attention mechanism computes attention scores between tokens using the dot product of Q and K, normalizes scores with softmax, and then applies them to V to generate contextualized representations.
The output then passes through a Feed-Forward Network (FFN), which applies non-linear transformations to refine token representations.
Multi-head attention (MHA)~\cite{cordonnier2021multiheadattentioncollaborateinstead} and grouped-query attention (GQA)~\cite{ainslie2023gqatraininggeneralizedmultiquery} extend this by allowing multiple query heads to attend to the same sets of key and value heads, which greatly saves memory consumption.

\MyPara{LLM inference.} 
LLM inference involves two main phases: \textit{prefill} and \textit{decode}. The prefill phase processes the initial input sequence (i.e., prompt) and generates the first output token. This phase is \textit{compute-intensive} because all tokens are processed in parallel.
After that, the decode phase generates output tokens in an \textit{auto-regressive} manner, generating one token at a time~\cite{vaswani2023attentionneed}. For each token, it computes a new query (Q) and performs self-attention over the key (K) and value (V) tensors of all previously generated tokens.
To avoid redundant computation, a \kv{} is employed to store the K and V tensors of past tokens in GPU memory. This significantly increases the usage of memory bandwidth, as each decoding step requires loading all stored KV tensors from memory, making the decode phase \textit{memory-intensive}~\cite{zhao2024atomlowbitquantizationefficient}.

\begin{figure}[!t]
    \centering
    \includegraphics[width=\linewidth]{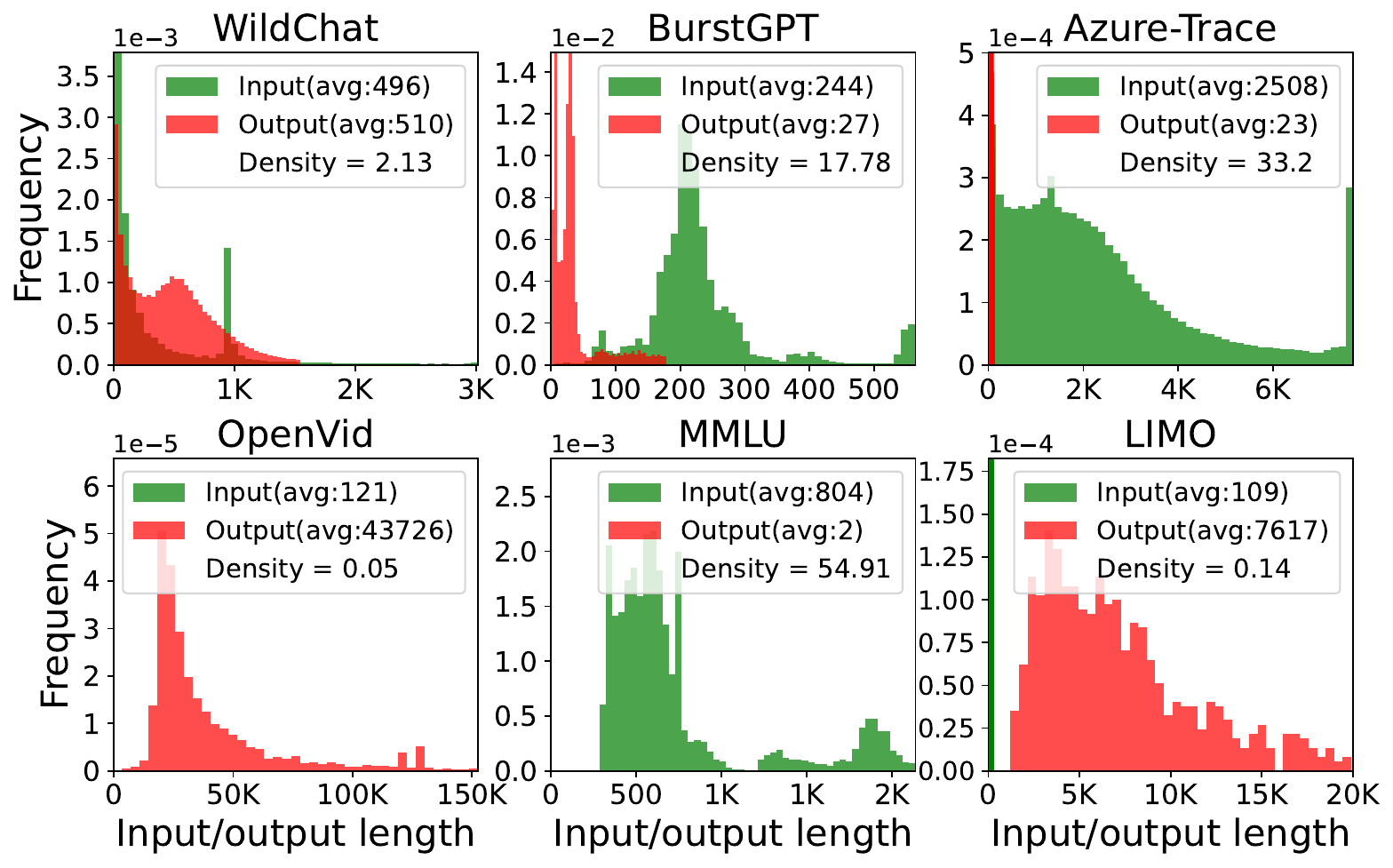}
    \caption{Request input/output length distribution from $6$ well-known open-sourced traces, including chatbot WildChat and API services BurstGPT~\cite{zhao2024wildchat1mchatgptinteraction,wang2024burstgpt}, Azure-Trace~\cite{stojkovic2024dynamollmdesigningllminference}, video generation datasets OpenVid~\cite{nan2024openvid1mlargescalehighqualitydataset}, benchmark traces MMLU~\cite{hendrycks2021measuringmassivemultitasklanguage} and math traces LIMO~\cite{ye2025limoreasoning}.
    Requests from different traces demonstrate distinct length distributions, which leads to different \density{}. Compute density is the ratio of compute to memory bandwidth usage (formally defined in \S\ref{sec-analysis}). A dataset is compute intensive when its compute density $>1$, and memory intensive otherwise.
    }
    \label{fig:motivation-traces-length}
\end{figure}

\subsection{Inference latency and throughput optimizations}
\label{sec:bg-opts}
Here, we introduce prior inference latency and throughput optimizations relevant to the design of \sys{}.

\MyPara{Prefill/Decode (P/D) disaggregation.}
\new{Early-stage inference systems use naive continuous batching scheduling~\citep{orca}, which overlooks the resource usage differences between prefill and decode phases. 
DistServe~\cite{zhong2024distservedisaggregatingprefilldecoding} proposes P/D disaggregation, which executes and scales these two phases independently on separate clusters. 
This allows time-to-first-token (TTFT) and time-per-output-token (TPOT) to be maintained independently without interference, making DistServe \emph{latency-optimized} for \emph{online} inference.
However, P/D disaggregation can reduce hardware utilization, making it suboptimal for throughput-oriented offline inference~\cite{drift,shi2025nexusproactiveintragpudisaggregationprefill,lin2025bulletboostinggpuutilization}. 
In particular, compute-intensive prefill phases saturate the compute resources of the prefill cluster while leaving memory bandwidth resources underutilized, and vice versa for the decode phase. We compare \sys{} with DistServe in \S\ref{sec-eval-subsec-e2e}.}

\MyPara{Phase-level colocation.}
\new{To solve this problem, Sarathi-Serve~\cite{agrawal2024tamingthroughputlatencytradeoffllm} proposed chunked prefill scheduling that colocates prefill and decode phases on the same clusters, and splits a large prefill into small chunks while adding only one chunk into the on-the-fly batch (i.e., requests currently being processed).}
Conceptually, chunked prefill achieves phase-level overlapping which uses both compute and memory resources, thereby improving arithmetic intensity per iteration and enhancing hardware utilization. 
However, chunked prefill was initially designed for online inference, where strict latency constraints prevent flexibly reordering requests to form a batch.
Therefore, when a set of requests consists mostly of memory-intensive requests, Sarathi-Serve will quickly run out of prefill phases, leaving GPU compute resources underutilized in the remaining decode processing.

\MyPara{Operator-level overlapping.}
Building upon P/D colocation (i.e., chunked prefill), a recent work, NanoFlow~\cite{zhu2024nanoflowoptimallargelanguage}, explores \emph{operator-level} resource overlapping. It splits a batch into micro-batches and overlaps compute-intensive GEMM operators with memory-intensive attention operators between micro-batches. 
\new{Another prior work, Orion~\cite{orion}, also explores operator-level GPU multiplexing by transparently scheduling distinct operators to maximize hardware utilization. 
This type of fine-grained overlapping is particularly beneficial when the batch contains a proper mix of prefill and decode tokens that can balance the execution time of GEMM and attention operators. However, both NanoFlow and Orion overlook the impact of request ordering on batch composition, limiting their ability to optimize throughput in \emph{offline} inference. For instance, if a workload begins with compute-intensive requests followed by memory-intensive ones, these frameworks process the batches sequentially rather than interleaving them, leading to suboptimal resource utilization. }

\MyPara{Prefix sharing.}
Prefix sharing (caching)~\cite{lin2024parrotefficientservingllmbased,zheng2024sglangefficientexecutionstructured, vllm} is a commonly adopted optimization that caches computed prompts from previously processed requests and reuses them for future requests. 
When a new request arrives, the system checks the cached prompts, and if a cache hit occurs, the shared prefix is reused, eliminating redundant computation and boosting throughput~\cite{cascade-inference}.
\new{
Prefix sharing provides considerable throughput gain for both compute- and memory-intensive workloads without hurting generation quality, e.g., studies show that certain workloads can save up to $80$\% computation~\cite{zheng2024sglangefficientexecutionstructured},
so it has been widely used in mainstream frameworks~\cite{vllm, deepseekContextCaching}. 
}
To enable efficient look-up, prefixes are organized using a Trie Tree~\cite{zheng2024sglangefficientexecutionstructured}, where each node is a segment of a prefix, and a complete path from the root to the leaf corresponds to a unique prefix. 
The prefix cache is stored alongside the regular \kv{} in GPU memory. 
When GPU memory runs out, the prefix cache may be evicted. Therefore, the access pattern can affect cache hit rates, which is denoted as \textit{\psr{}} in this work.

\section{Motivation}

\subsection{Evolving workloads diversity}
\label{sec:request-diversity}

The capabilities of LLMs are evolving rapidly.
First, multi-modality advancements have enabled modern models (e.g., LWM\citep{liu2024worldmodelmillionlengthvideo}, Unified-IO\citep{lu2023unifiedio2scalingautoregressive}, EMU\citep{wang2024emu3nexttokenpredictionneed}, MIO\citep{wang2024miofoundationmodelmultimodal}, and VILA-U~\citep{wu2024vilauunifiedfoundationmodel}) to process diverse input and output modalities, including text, images, videos, and their combinations.
These models typically share a common architecture: a transformer-based LLM augmented with modality-specific adapters. 
These adapters convert inputs from various modalities into a format that the base model can process and translate its outputs back into the desired modality. 
In addition, the emergence of reasoning models enables models to ``think'' before generating answers~\cite{snell2024scalingllmtesttimecompute,qin2024o1replicationjourneystrategic,yao2023treethoughtsdeliberateproblem,wei2023chainofthoughtpromptingelicitsreasoning,tang2024questqueryawaresparsityefficient}, which greatly improves their performance on hard tasks such as math and coding. 

As a result, LLM-based applications are expanding rapidly, exhibiting increasing \textit{workload diversity}, i.e., diverse input and output token lengths. 
To visualize this diversity, we present the request length distributions in different use cases in Figure~\ref{fig:motivation-traces-length}\footnote{Some traces are collected from online inference, but similar distributions can be observed in offline inference. For example, text-based chat and benchmarks are commonly used for model evaluation using offline processing; video generation can be leveraged to produce game summaries offline.}.
It shows that text-only chat requests typically have hundreds of tokens but a video generation request can easily generate tens of thousands tokens.
While the simple questions in the MMLU benchmark produce only a few tokens, hard questions from the LIMO benchmark can produce thousands of tokens.

\subsection{Workload diversity limits existing overlapping}
\label{sec:request-resource-imbalance}

\MyPara{Diverse resource demands across requests.}
These diverse requests consume GPU resources (i.e., compute and memory bandwidth) differently. 
Since prefill is compute-intensive, requests with long inputs but short outputs will consume more GPU compute than memory bandwidth. Conversely, requests with long \outlen{} use more GPU memory bandwidth due to their long memory-intensive decode phase.
Therefore, different request length distributions lead to drastically diverse resource demands across datasets. As formally defined in \S\ref{sec-analysis}, we use compute density to represent the ratio of compute to memory bandwidth usage, with higher values indicating more compute-intensive.
As shown in Figure~\ref{fig:motivation-traces-length}, OpenVid~\cite{nan2024openvid1mlargescalehighqualitydataset} and LIMO~\cite{ye2025limoreasoning} are highly memory-intensive while the remaining datasets are more compute-intensive.

\begin{figure}[!t]
    \centering
    \includegraphics[width=0.9\linewidth]{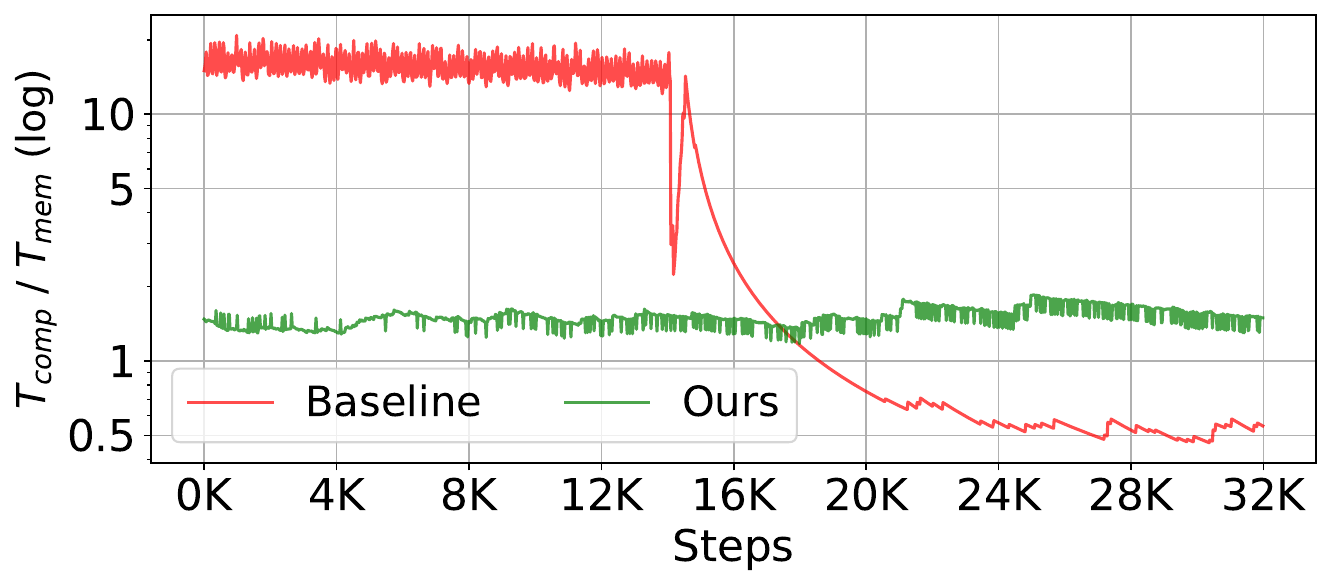}
    \caption{
    The ratio of time spent on compute-bound and memory-bound operations, when serving \lm{}-3-8B on an A100 GPU. The workloads are synthesized by sequentially combining compute-intensive (BurstGPT) and memory-intensive (OpenVid) traces. The baseline causes underutilization of one resource at each execution step, while ours achieves stable and balanced resource usage.
    }
    \label{fig:motivation-utilization}
\end{figure}

\MyPara{Intra-batch optimizations alone are insufficient.}
This request diversity presents significant challenges in maximizing inference throughput.
Prior studies, such as chunked prefill~\citep{agrawal2024tamingthroughputlatencytradeoffllm}, Orion~\cite{orion}, and NanoFlow~\citep{zhu2024nanoflowoptimallargelanguage}, optimize throughput by overlapping compute and memory bandwidth usage within a batch of requests. \new{For example, chunked prefill colocates the prefill phase with the decode phase within the same batch to overlap compute and memory usage.}
However, without considering the resource demands across batches, their effectiveness diminishes when a batch is dominated by either compute-intensive or memory-intensive requests, as the system can be easily bottlenecked by one type of resource while leaving the other underutilized.

To illustrate this, we compare NanoFlow (state-of-the-art throughput-oriented system) against our system by measuring the total time spent on compute- and memory-bound operators when serving a workload with compute-intensive requests in front followed by memory-intensive requests. 
As shown in Figure~\ref{fig:motivation-utilization}, NanoFlow serves requests sequentially, underutilizing memory bandwidth when processing compute-intensive requests and compute resources during processing memory-intensive requests. In contrast, our system strategically reorders requests with complementary resource demands, resulting in balanced resource utilization and increased overall throughput.

\subsection{Resource-aware batching via request reordering}
\label{sec:formulate-optimal-throughput}

The above problem has motivated us to consider the diverse resource demands when batching requests.
Our key idea is to exploit the \textit{relaxed latency constraints} of offline inference to reorder requests and create batches that can maximize the benefit of compute-memory overlapping, improving GPU utilization and increasing throughput.

\MyPara{Challenge: conflicts between resource overlapping and prefix sharing.}
\new{However, resource overlapping can conflict with prefix sharing, a widely used technique that significantly improves throughput by saving redundant computation~\cite{lin2024parrotefficientservingllmbased,zheng2024sglangefficientexecutionstructured}. }
As introduced in \S\ref{sec:bg-opts}, inference systems structure the prefix cache with a Trie Tree~\cite{zheng2024sglangefficientexecutionstructured}.
As proven in previous studies~\cite{zheng2024sglangefficientexecutionstructured,srivatsa2024prebleefficientdistributedprompt}, the request order that maximizes prefix sharing is to traverse the tree using Depth-First Search (DFS), ensuring that all shared prefixes are computed only once.
However, this order can conflict with the reordering needed to maximize resource overlap, leading to imbalanced resource demands within a batch, which in turn causes hardware underutilization and limited throughput. For example, when serving \lm{}-3-8B with one A100 GPU, DFS ordering can only achieve $71.7$\% of the optimal throughput, which maximizes both resource overlapping and prefix sharing (\refsec{sec-eval-subsec-e2e}), leaving a huge performance gap.

\MyPara{Our goal: harmonizing both for throughput optimization.} 
As a result, we must consider resource overlap and prefix sharing simultaneously to achieve the best of both. We formulate this problem as follows:
$$
T=f((1-s)\cdot T_{\text{comp}}, T_{\text{mem}})
$$
where $T$ is the total execution time of all requests, and $T_{comp}$ and $T_{mem}$ denote the total execution time of compute-bound and memory-bound operations across all requests, respectively. Detailed calculations of them will be provided in \S\ref{sec-analysis}; here, we focus on conveying the high-level formulation.
$s\in[0,1]$ here represents \psr{}, which means $s$ of the $T_{comp}$ are saved, so the compute time will be reduced to $(1-s)\cdot T_{\text{comp}}$. However, prefix cache hits do not reduce memory bandwidth usage, as the \kv{} still needs to be retrieved from memory.
$f$ is a function that depends on the scheduling policy and the request order. For example, for a policy that sequentially executes compute-bound and memory-bound operators (e.g., first-come-first-serve in \cite{vllm,zheng2024sglangefficientexecutionstructured}), \new{$f$ will be $sum(\cdot,\cdot)$ since compute and memory resources are utilized sequentially.}

\new{To minimize the end-to-end execution time $T$ to achieve optimal $T_o$, a perfect request scheduling is necessary to leave only the bottlenecked resource on the critical path while overlapping the other resources, namely $f = max(\cdot,\cdot)$. At the same time, all shared prefixes should be cached by prefix sharing without incurring any redundant computation, achieving an optimal prefix ratio $s_o$ which is determined by the workload prompts. In the rest of this paper, we will describe how \sys{} approaches $T_o$ through its design. }
$$
T_o=\max((1-s_o)\cdot T_{\text{comp}}, T_{\text{mem}})
$$

\section{Performance Analysis}
\label{sec-analysis}

In this section, we formally define \textit{compute density}, \new{a metric that quantifies the ratio of compute and memory resource usages}. This metric enables \sys to analyze diverse resource demands across requests and guides its scheduling to balance compute and memory usage for effective overlapping. Besides, compute density provides a practical method to approximate $T_o$.

\subsection{Request-level compute density}
\label{sec-analysis-request-density}
We first define compute density at the request level and extend it to the batch level in \S\ref{sec-req-level-to-batch-level}.
We define the \density{} $\rho(r)$ of a request $r$ as the total compute time of compute-intensive operators divided by the total time of memory-intensive operators, following the similar intuition of arithmetic intensity~\cite{roofline}:
$$
\rho(r)=\frac{\text{Comp}(r)}{\text{Mem}(r)}
$$
\new{where a larger compute density $\rho(r)$ indicates a request that requires more compute resources rather than memory bandwidth (i.e., compute-intensive). Note that the following formulations assume an unquantized data type, FP16, as well as GPU tensor core computation capability. One can easily adapt the data type and GPU capability by varying the constants in the formulas.}

Next, to calculate $\rho(r)$, we build a resource usage model for a request with \inlen{} $p$ and \outlen{} $d$. 
\new{Input length of a request is known as the prompt length, and we will discuss how to estimate the output length in \S\ref{sec:design:data-structure}.}
Given a model of $P_{model}$ parameters, $H$ hidden dimension of model width, $H_{kv}$ feature dimension for each KV head, and $L$ decoder layers, and a hardware configuration of $\texttt{compute}$ peak FP16 GFlops and $\texttt{bandwidth}$ GB/s memory bandwidth, the total time for compute-bound operators of a single request $r$ can be approximated by total computation amount of GEMM operators and the self-attention in prefill phase divided by the hardware compute capability:
$$
\text{Comp}(r) \approx \frac{2 \cdot (p + d) \cdot P_{model}+4\cdot p^2\cdot H\cdot L}{\texttt{compute}}
$$
where $(p+d)$ is the number of tokens processed by GEMM operators during the lifetime of $r$. Since parameters of GEMM ($QKV$ generation + FFN) occupy most of the model parameters, the computation amount can be effectively approximated by the $\texttt{model\_size}$, $P_{model}$~\cite{zhu2024nanoflowoptimallargelanguage}.
\new{ Since the attention consists of $2$ GEMMs including $P=Q\times K$ and $P\times V$ where each GEMM leads to $2\cdot p\cdot p\cdot H$ Flops, the total computation amount is then multipied with $L$ layers, i.e., $4\cdot p^2\cdot H\cdot L$.
The $p^2$ comes from the quadratic computation of self-attention in the prefill phase. As $p\cdot H\cdot L$ is typically much smaller than $P_{model}$ on common workloads with $p$ of a few hundred tokens (Figure~\ref{fig:motivation-traces-length}), we omit $4\cdot p^2\cdot H\cdot L$ in the following deduction.}

The total time for memory-bound operators can be approximated by counting the total memory loading of $d$ times decoding attention during the auto-regressive generation:
\begin{align}
\text{Mem}(r) &\approx \frac{\sum_{i=1}^{d}(p+i)\cdot H_{kv} \cdot L \cdot 2 \cdot 2}{\texttt{bandwidth}} \notag \\
&= \frac{(p \cdot d + \frac{1}{2} \cdot d^2) \cdot H_{kv} \cdot L \cdot 4}{\texttt{bandwidth}} \notag
\end{align}
where $\sum_{i=1}^d(p+i)$ calculates the total number of loaded tokens by self-attention during the $d$ steps of the auto-regressive generation process, and $4$ comes from key and value vectors stored in FP16 for each token.

\subsection{Translating request-level metrics to batch-level}
\label{sec-req-level-to-batch-level}

Ideally, a scheduling policy should reorder requests to form batches with perfectly balanced $T_{\text{comp}}$ and $T_{\text{mem}}$. However, achieving this balance is difficult using only a \textit{request-level \density{}} metric, 
\new{
as requests in the same batch may reside in different inference steps that affect $T_{\text{comp}}$ and $T_{\text{mem}}$ differently. 
}
For example, adding a memory-intensive request does not immediately lower a batch's overall \density{}, because the request will undergo a compute-intensive prefill phase first, only becoming memory-intensive later during its decode phase. Therefore, measuring only the \density{} of individual requests is insufficient. Instead, \sys{} must consider each request's compute intensity across its entire generation lifetime, requiring a \textit{holistic batch-level metric}.

Fortunately, integrated with continuous batching~\cite{orca}, a batch typically consists of many requests in different steps, and request-level \density{} essentially captures the average compute intensity over time, making it a good approximation for the \density{} of a batch. \new{Specifically, when the requests within the batch are evenly distributed at different steps, batch-level \density{} will converge to request-level \density{} for requests with \inlen{} of $p$ and \outlen{} of $d$. We demonstrate this following the same notations in \refsec{sec-analysis-request-density}.}

Denoting the total memory capacity of \kv{} as \texttt{KV-Mem}, we can calculate batch-level \density{} with the total compute time and memory loading time. Since a batch typically consists of a large number of tokens, $\text{Comp}(B)$ is dominated by GEMM computation, and $\text{Mem}(B)$ is dominated by loading of \kv{}, \new{compared to the small operators including layer normalization, activation, and position embedding}. Therefore, we have: 

\begin{align*}
    \text{Comp}(B) \approx \frac{\frac{\texttt{KV-Mem}}{(p+\frac{d}{2})\cdot H_{kv}\cdot L\cdot 4}\cdot\frac{p+d}{d}\cdot P_{model}\cdot 2}{\texttt{compute}}
\end{align*}
where the average length of \kv{} per request is $p+\frac{d}{2}$, and the number of decoding requests $B_{decode}$ is $\texttt{KV-Mem}$ divided by $(p+\frac{d}{2})$ tokens.
\new{As each token takes $H_{kv}\cdot L\cdot 4$ bytes, $B_{decode}$ can be calculated as $\frac{\texttt{KV-Mem}}{(p+\frac{d}{2})\cdot H_{kv}\cdot L\cdot 4}$. As chunked-prefill scheduling maintains a stable batch size, the number of average newly admitted requests should be equal to the average completed requests, which indicates that the ratio of prefill tokens with decode tokens is $\frac{p}{d}$. Therefore, the prefill tokens can be calculated as $B_{decode} \cdot\frac{p}{d}$, leading to a total number of tokens as $B_{decode}\cdot\frac{p+d}{d}$. As discussed in~\refsec{sec-analysis-request-density}, each token contributes to a total amount compute of $2\cdot P_{model}$, which concludes the $\texttt{Comp}(B)$ by substitution.
}

The total loading time of \kv{} within a batch $B$ is:

\begin{align*}
    \text{Mem}(B) \approx \frac{\texttt{KV-Mem}}{\texttt{bandwidth}}
\end{align*}

We show the equivalence of batch-level \density{} $\rho(B)$ and request-level \density{} $\rho(r)$ with the following derivation:

\resizebox{0.95\linewidth}{!}{
\begin{minipage}{\linewidth} %
\begin{align*}
\mathlarger{\bm{\rho(B)}} & \bm{=} \frac{\text{Comp}(B)}{\text{Mem}(B)}
        \approx \frac{\frac{\texttt{KV-Mem}}{(p+\frac{d}{2})\cdot H_{kv}\cdot L\cdot 4}\cdot\frac{p+d}{d}\cdot P_{model}\cdot 2}{\texttt{compute}} \bigg/ \frac{\texttt{KV-Mem}}{\texttt{bandwidth}}\\
        & = \frac{(p+d)\cdot P_{model}\cdot2}{\texttt{compute}} \bigg/ \frac{(p+\frac{1}{2}\cdot d)\cdot d\cdot H_{kv}\cdot L\cdot 4}{\texttt{bandwidth}} %
        \approx \mathlarger{\bm{\rho(r)}}
        \label{eq-batch-density}
\end{align*}
\end{minipage}
}

Such derivation of batch-level \density{} can also be cross-validated with previous literature~\cite {zhu2024nanoflowoptimallargelanguage}. Therefore, \sys{} adopts request-level \density{} as the key metric to make scheduling decisions and is still able to accurately control batch-level \density{} for efficient resource overlapping.

\subsection{Case study: offline inference with Llama-3-70B}
\label{sec-case-study}
\begin{figure}[!t]
    \centering
    \includegraphics[width=0.95\linewidth]{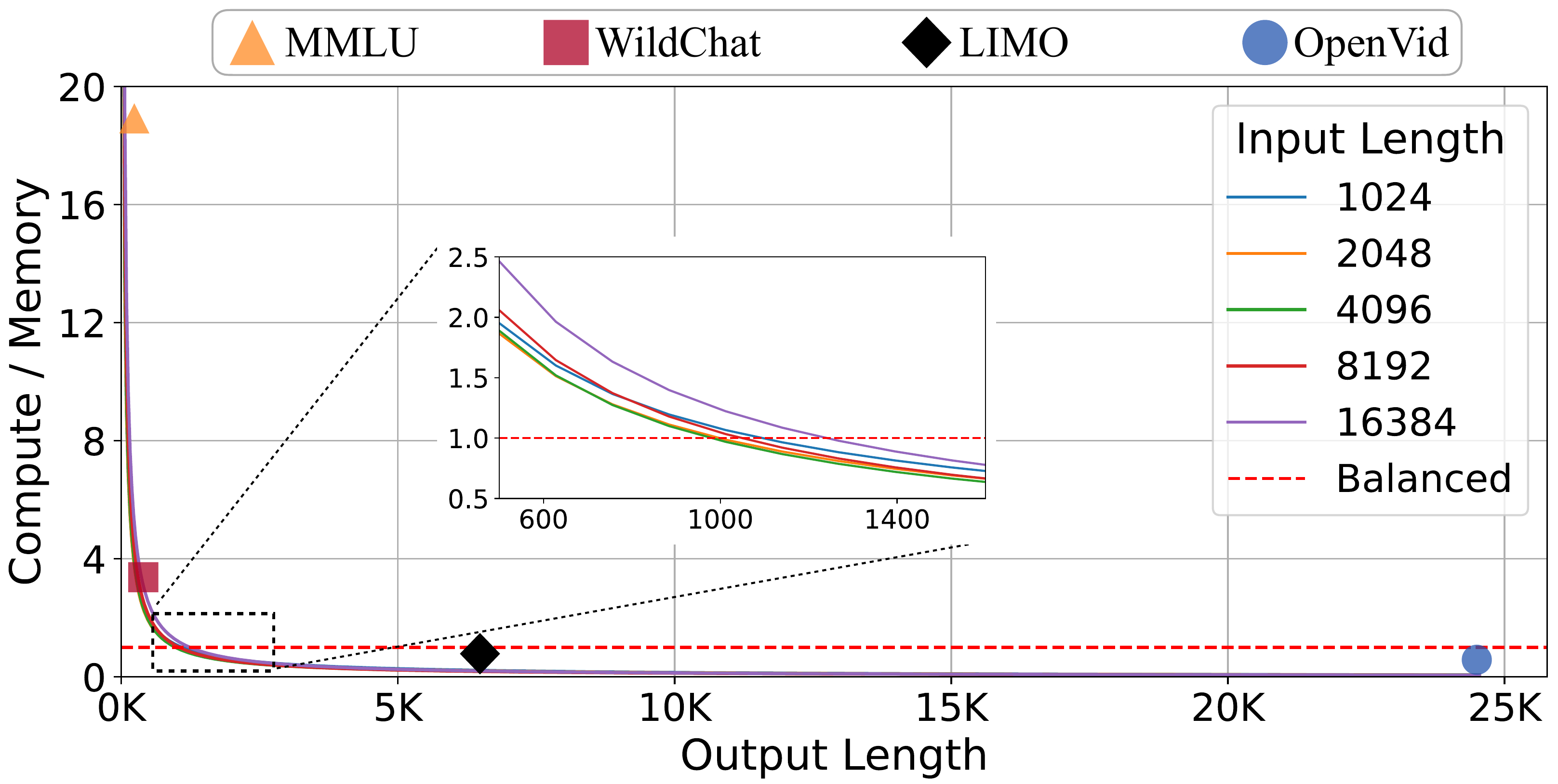}
    \caption{Compute density of requests with different input/output lengths (\lm-3-8B on an A100 80GB GPU) varies drastically and leads to diverse resource demands.
    }
    \label{fig:motivation-compute-density}
\end{figure}

\begin{table}[t]
\resizebox{\linewidth}{!}{
    \centering
    \begin{tabular}{|l|c|c|c|}
        \hline
        Time (ms) & Batch size=512 & Batch size=768 & Batch size=1024 \\
        \hline
        GEMM & 1.038 / 1.087 & 1.494 / 1.537 & 1.916 / 2.005 \\
        \hline
        Attention & 1.239 / 1.317 & 1.859 / 1.913 & 2.478 / 2.515 \\
        \hline
    \end{tabular}
}
    \caption{Operator performance differences for varying batch sizes with a sequence size of 1024 (estimated time / real execution time).}
    \label{tab:kernel-exec-time}
    \vspace{-8mm}
\end{table}

To visualize the drastic differences in compute density across datasets and validate the accuracy of our performance model, we conducted a case study using Llama-3-8B on an A100 80GB GPU and requests with varying \inlen{} $p$ and \outlen{} $d$.
As shown in Figure~\ref{fig:motivation-compute-density}, compute density diminishes quickly for requests with longer \outlen{}, indicating their memory-intensive nature, as exemplified by OpenVid~\cite{nan2024openvid1mlargescalehighqualitydataset}. In contrast, requests from WildChat~\cite{zhao2024wildchat1mchatgptinteraction} and MMLU~\cite{hendrycks2021measuringmassivemultitasklanguage} typically have short \outlen{}s and remain compute-intensive.

To further validate our performance model proposed in \S\ref{sec-analysis-request-density}, we compare its estimated times against measured execution times in Table~\ref{tab:kernel-exec-time}. The estimated times closely match actual execution times for both GEMM and attention kernels, with a maximum relative error of 6\%.

\section{\sys{} Design}
\label{sec-design}

\begin{figure*}[!t]
    \centering
    \includegraphics[width=0.80\linewidth]{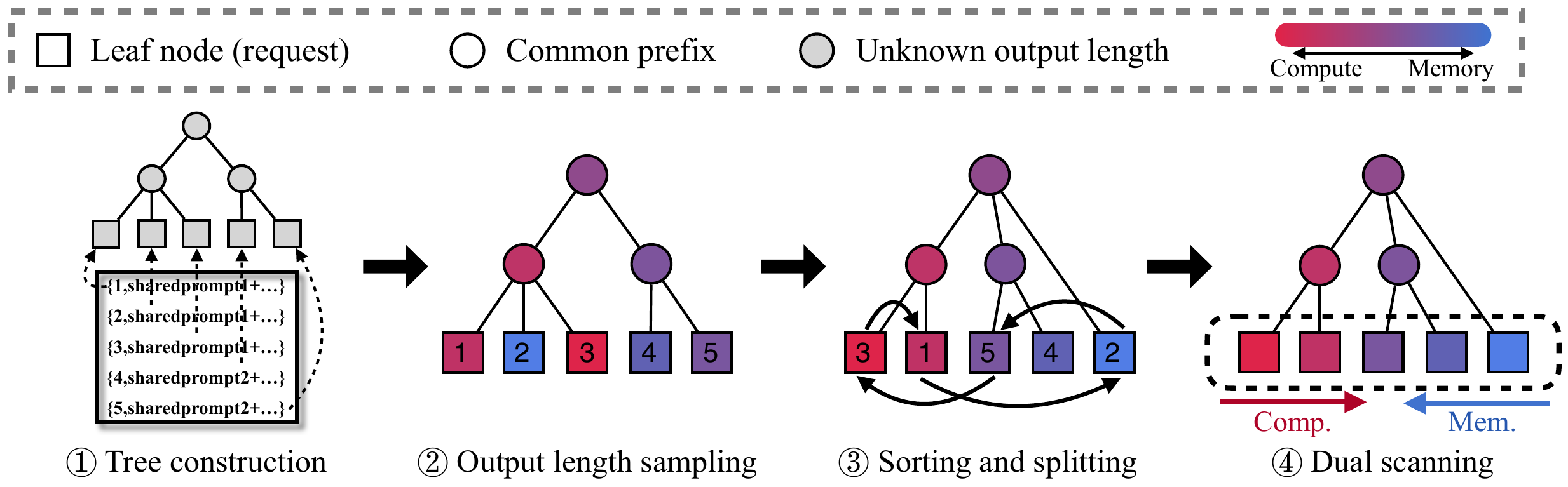}
    \caption{Overview of \sys{}'s design. Leaf nodes in the prefix tree are actual requests while others represent the shared prefix in prompts. The color of nodes represents the resource demand of all requests within the sub-tree, which is more compute-intensive in red and memory-intensive in blue. Given a set of requests, a one-time warm-up ahead of GPU running is performed, which consists of prefix tree construction, \outlen{} sampling, and transformation including tree sorting and node splitting (\ding{192},\ding{193}, and \ding{194}). Then the dual scanner forms the runtime batch from most compute- and memory-intensive nodes, which is consumed by the backend engine (\ding{195}). This warm-up is a short process and finishes quickly within the first $1$\% of time during the end-to-end inference generation.
    }
    \label{fig:design-overview}
\end{figure*}

\textbf{Overview.}
Figure~\ref{fig:design-overview} shows the end-to-end workflow of \sys{}. 
Given a set of requests upfront with known prompts, \sys{} first constructs a prefix tree to capture the shared prefix among requests (\ding{192}, \refsec{sec:design:data-structure}).
Next, \sys{} calculates compute density for each node, which involves estimating request \outlen{} by sampling over the prefix tree (\ding{193}, \refsec{sec:design:data-structure}). 
With compute density, requests are characterized as compute- or memory-intensive and sorted based on their resource usage, resulting in a sorted tree where most compute-intensive requests are on the left and most memory-intensive requests are on the right (\ding{194}, \refsec{sec:design:transformation}). 
Therefore, \sys{} can efficiently find a request order by sweeping the tree from left and right simultaneously. This order can balance compute-memory demand for resource overlapping and has high prefix sharing (\ding{195}, \refsec{sec:design:dual-scanner}).
Finally, the ordered requests are batched and fed into a backend engine for inference. 
To support large-scale deployment with more GPUs, \sys{} integrates both data and tensor parallelism (\refsec{sec:design:distribute}).

\subsection{Key data structure: resource-aware prefix tree}
\label{sec:design:data-structure}
Determining the optimal scheduling order requires a proper abstraction that can capture both shared prefixes and resource demands of all requests.
Inspired by the Trie Tree data structure in \textit{RadixAttention}~\cite{zheng2024sglangefficientexecutionstructured}, \sys{} organizes all requests within a \textit{resource-aware prefix tree}, where each leaf node represents an actual request and each internal node is a segment of the prefix shared by all its descendants.
Therefore, a path from the root node to the leaf node represents the longest shared prefix of this request.
By traversing this prefix tree in a DFS order, each internal node (i.e., shared prompt segment) is visited with the shortest reuse distance, which gives a request sequence that maximizes the \psr{}~\cite{zheng2024sglangefficientexecutionstructured}. However, such naive DFS ordering neglects diverse resource demands across requests and misses the opportunity for resource overlapping.

To harmonize prefix sharing and resource overlap, we enhance the prefix tree with resource demand information for each node, making it a resource-aware prefix tree.
Specifically, we compute the \textit{compute density for each node} by considering its prefix sharing status, as defined below:
$$
\rho(R)=\frac{(1-s)\cdot T_{\text{comp}}}{T_{\text{mem}}}
\vspace{-1mm}
$$
where $R$ represents the set of requests in the node, and $s$ denotes its \psr. For an internal node of the tree, the \density{} is calculated over all requests within the sub-tree rooted at it.
With this enhancement, the resource-aware prefix tree provides a \textit{unified abstraction} that enables \sys to efficiently search for the optimal request order that harmonizes both prefix sharing and resource overlap.

\MyPara{\Outlen{} sampling.}
Request \outlen{} is necessary for calculating compute density as modeled in \refsec{sec-analysis-request-density}, which is \textit{unknown} beforehand because LLMs generate tokens in an auto-regressive manner.
As a result, an estimation mechanism before actual inference is needed. 
\new{Our observation here is that a request's \textit{\outlen{} distribution} is closely related to its \textit{prompt semantics} and \textit{task type}~\cite{fu2024efficientllmschedulinglearning,stojkovic2024dynamollmdesigningllminference,arango2025prefix,zheng2025batchllmoptimizinglargebatched}.} 
For example, benchmark requests (e.g., MMLU~\cite{hendrycks2021measuringmassivemultitasklanguage}, LongBench~\cite{bai2024longbenchbilingualmultitaskbenchmark}) have an \outlen{} of only a few tokens~\cite{hendrycks2021measuringmassivemultitasklanguage}, while chatbot (e.g., ShareGPT~\cite{sharegpt}, WildChat~\cite{zhao2024wildchat1mchatgptinteraction}) generates an average of hundreds of tokens~\cite{chen2023punicamultitenantloraserving}.

Such an observation unveils a unique opportunity in \offinf{}, where a batch of requests submitted by a user typically shares the same task type or shared prefixes. In the prefix tree, requests sharing similar prompts are naturally grouped under specific sub-trees. Therefore, these requests tend to have a similar distribution of \outlen{}. 
To estimate \outlen{}, \sys{} selects a subset of requests with a sampling probability $p$ to undergo the full inference process and obtain their \outlen{} in the warmup phase.
Each sub-tree uses the average \outlen{} of its sampled requests as an estimation for the left unsampled requests within the same sub-tree. 
If a sub-tree $t_1$ is not sampled at all, it will use the average sampled \outlen{} of its sibling sub-tree $t_2$ since $t_1$ and $t_2$ share the longest common prefix and tend to have a similar distribution of \outlen{}.
This sampling process does not incur any \textit{extra overhead} as sampled requests can be directly returned to the user without running inference again.

\begin{figure}[!t]
    \centering
    \includegraphics[width=0.7\linewidth]{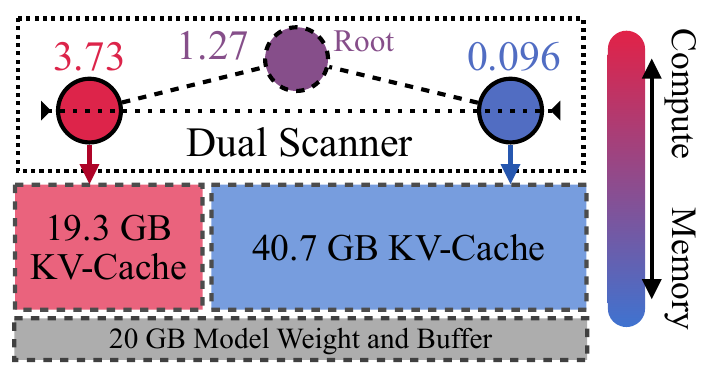}
    \caption{
    An example of \sys{}'s memory partition with 80GB memory. The left node has a \density{} of $3.73$, while the right node is memory-intensive with a \density{} of $0.096$. \new{The dual scanner will reserve $20$GB for model weights and temporary buffers, then partition the rest $60$GB to reach the root density of $1.27$. Given the compute densities, the memory is partitioned into $19.3$ and $40.7$GB , where $3.73\times19.3+0.096\times40.7=1.27\times 60$.}
    }
    \label{fig:design-memory}
    \vspace{-2pt}
\end{figure}

\subsection{Resource-aware prefix tree sorting}
\label{sec:design:transformation}
Next, \sys{} performs a layer-wise sorting of nodes based on their \density{}, which only reorders nodes sharing the same ancestor and depth (detailed algorithm in \refsec{sec:appendix-pseudo-algorithm}). Therefore, this sorting maintains the hierarchical structure of the prefix tree.
After sorting, the tree exhibits a global pattern with compute-intensive nodes on the left and memory-intensive nodes on the right. 
However, local outliers that deviate from this trend may still exist. For instance, in the first tree of Figure~\ref{fig:design-overview}, request \#2, which has low compute density, should be separated from requests \#1 and \#3 and repositioned to the right.
            
To address this issue, \sys{} introduces a conditional node splitting technique to relocate outliers to desired positions (detailed algorithm in \refsec{sec:appendix-pseudo-algorithm}). The node that is split from the original node will be inserted at the root when there is no shared prefix at the desired position, potentially incurring prefix recomputation costs during inference. Additionally, the compute density of the original node, the split node, and the new parent need to be updated accordingly. Take Figure~\ref{fig:design-overview} \circled{\small{3}} as an example, request \#2 is moved from the leftmost to the rightmost position, requiring its prefix to be recomputed. This technique applies a heuristic threshold $t$: if the recomputation overhead for relocation falls below $t$, the node is repositioned to preserve the descending order of compute density. This approach enables a controlled trade-off, sacrificing a small degree of prefix sharing to better order requests with their resource demands for \sys's request scheduling. In practice, we found \sys{}'s performance is insensitive to $t$ for real-world workloads (discussed shortly in \refsec{sec:design:robustness-analysis}) and \sys{} works generally well when we set it to preserve 99\% of prefix sharing ratio.

\subsection{Request order search: heuristic dual scanning}
\label{sec:design:dual-scanner}

Finally, \sys{} derives a request order for batching, with the aim of achieving both high \psr{} and resource overlap across inference iterations.

Searching for an optimal request order is NP-hard. For each scheduling step, the search problem can be reduced to a knapsack problem~\cite{knapsack} where requests with different KV cache sizes (cost) and compute density values (value) are selected to fill the GPU memory for the targeted density score.
Furthermore, since requests undergo multi-step decoding in auto-regressive inference, scheduling in different steps is dependent, further complicating the problem. 
Given the large number of requests and scheduling steps, finding the optimal solution in a reasonable time is infeasible.

To solve this problem in a reasonable time, \sys{} employs a \textit{heuristic-based} algorithm that scans the leaf nodes of the prefix tree concurrently from left to right and right to left, progressively adding requests to the on-the-fly batch during this process. 
\new{By controlling the ratio of the number of requests admitted from these two ends, an arbitrary and stable compute density can be achieved, thus improving the resource balance. To determine how many requests should be selected from the current compute-intensive node $R_L$ and memory-intensive node $R_R$, \sys{} first calculates the desired memory capacity for each side and then adds requests to saturate the assigned memory.} \sys{} logically partitions the GPU memory $M$ into two parts $M_L$ and $M_R$, where the partition sizes $M_L$ and $M_R$ are dynamically calculated by the following theoretical constraints:
\begin{equation*}
\left\{
\begin{array}{ll}
    M_L + M_R = M & \text{(Memory)} \\
    M_L \cdot \rho(R_L) + M_R \cdot \rho(R_R) = M \cdot \rho(rt) & \text{(Compute)}
\end{array}
\right.
\end{equation*}
These two equations represent the memory and compute demands, respectively. Here, $M$ is a constant denoting GPU memory size.
$\rho(rt)$ is the compute density of the tree root node, which remains as a constant for the current request set.
\new{Similarly, $\rho(R_L)$ and $\rho(R_R)$ are the compute densities of the compute- and memory-intensive nodes, which are also constants when the scanner reaches a specific node. 
Given these constants, the first equation limits the total memory allocation to the available GPU memory, while the second equation constrains the total compute to match the target density $\rho(rt)$. Together, these two constraints achieve $\rho(rt)$ by combining requests with densities $\rho(R_L)$ and $\rho(R_R)$. }
Thus, $M_L$ and $M_R$ can be derived from these two equations. We illustrate one practical example in Figure~\ref{fig:design-memory}.

Given an assigned memory size, \sys can calculate the desired on-the-fly batch size and construct the batch by selecting requests from $\rho(R_L)$ and $\rho(R_L)$ accordingly, ensuring that they can be placed into $M_L$ and $M_R$ respectively.
This memory partition ensures that the compute density of the blended compute- and memory-intensive requests approximates $\rho(rt)$, allowing the memory access time to be fully overlapped with the compute time (when $\rho(rt)>1$). 
Moreover, this strategy also ensures high \psr{}, as the dual scanning method essentially traverses the prefix tree in DFS order from both sides. \new{We include the detailed algorithm of dual scanning in~\refsec{sec:appendix-pseudo-algorithm} (Algorithm~\ref{alg:dual-scan}).}

\subsection{Robustness analysis}
\label{sec:design:robustness-analysis}

\MyPara{Handling inaccurate output length estimation.}
Notably, predicting \outlen{} may not always be accurate due to the dynamic nature of decoding, except for image- or video-generation, where \outlen{} is inherently predefined by the preset quality and frame parameters~\citep{liu2024worldmodelmillionlengthvideo,lu2022unifiediounifiedmodelvision}. 
Fortunately, \sys{} does not require precise \outlen{} predictions due to the following reasons.
First, a rough estimation sufficient to distinguish request types (e.g., benchmark v.s. conversational tasks) is adequate for \sys. This is because \sys{} processes hundreds of requests in a single batch to overlap compute and memory, minor estimation deviations within the same request type have negligible impact on overall batch performance. To verify this, we only sampled $1$\% of the total requests for \outlen{} sampling and found that \sys{} can achieve comparable end-to-end performance to a sampling probability of $100$\%. 
In addition, \sys can online adaptively adjust the batch to mitigate the impact of miss-estimations. If a request finishes much earlier due to an overestimated \outlen{}, \sys{} will insert additional requests. Conversely, if \outlen{} is severely underestimated, \sys{} could relocate the request from $M_L$ into $M_R$.

\MyPara{Stopping conditions and convergence.}
\new{
The algorithm iteratively performs
``layer-wise sort $\to$ conditional node split $\to$ (re)sort'' until one of the following holds: (C1) the leaf sequence ordered by compute density becomes non-increasing, or (C2) for every leaf, the split cost exceeds the threshold $t$.
Therefore, termination is guaranteed: after each split, the produced leaf is reinserted as a direct child of the root. In the worst case, every original leaf is split once and moved under the root; a single layer-wise sort at the root then yields a globally monotone order, satisfying (C1).
Since the number of original leaves is finite, each leaf can be split at most once, so the total number of splits is $\leq N_{\text{leaf}}$ and the number of  (re)sorts is $\leq N_{\text{leaf}} + 1$.
Empirically, due to the threshold $t$, only about 0.1\% to 1\% of leaves require splitting. By tuning $t$ we obtain a controllable performance bound.}

\MyPara{Performance robustness of tree sorting.}
Since the optimal ordering for prefix sharing and resource overlapping can sometimes conflict, our tree sorting and node-splitting algorithm may perform differently depending on workload characteristics. 
However, real-world workloads typically expose low variance in request compute density within each dataset, thus delivering near-optimal performance.

\subsection{Distributed deployment}
\label{sec:design:distribute}

\sys{} supports \textit{data parallelism} and \textit{tensor parallelism} for efficient scaling across different number of GPUs.

\MyPara{Data parallelism.}
Data parallelism (DP) extends computational capacity by distributing identical model replicas across hardware clusters, each performing computations on distinct subsets of data with identical control flows. To implement DP effectively, \sys{} first constructs the centralized resource-aware prefix tree with the entire request pool, and then decomposes it into \textit{parallelized subtrees} assigned to different DP ranks. Such decomposition ensures balanced workloads and resource usage across partitions. \sys{} reuses the dual-scanner design to form request partitions as subtrees. Once a subtree reaches the target workload, \sys{} finalizes it and starts a new one.
This approach incurs only marginal prefix sharing overhead due to tree partitioning---one path from the tree root to the leaf cannot be shared across DP replicas, but the impact is negligible.

\MyPara{Tensor Parallelism.}
Tensor parallelism (TP) partitions model parameters across multiple GPUs, addressing scenarios where a single GPU cannot accommodate the entire large model~\cite{shoeybi2020megatronlmtrainingmultibillionparameter}. 
Prior research has shown that the network communication overhead incurred by TP can be effectively overlapped through specialized pipeline strategies~\cite{zhu2024nanoflowoptimallargelanguage, centauri}. 
\sys is compatible with these designs, so it can seamlessly integrate TP with minimal performance degradation.

\section{Implementation and Evaluation}
\label{sec:eval}

\subsection{Implementation}
\label{sec:implementation}

We developed the resource-aware prefix tree based on SGLang \cite{zheng2024sglangefficientexecutionstructured} and enhanced it with node sorting and splitting driven by compute density.
Our scheduler is implemented based on NanoFlow~\cite{zhu2024nanoflowoptimallargelanguage}, which incorporates chunked prefill and continuous batching to improve system performance~\cite{agrawal2024tamingthroughputlatencytradeoffllm,orca}.
Our backend engine is built in C++ following NanoFlow's operator-level overlapping. It enables the simultaneous execution of compute-intensive operators like GEMM and memory-intensive operators like self-attention. \new{We include more implementation details in~\refsec{sec:appendix:impl-details}.}

\subsection{Experiment setup}
\label{sec:eval-setup}
\MyPara{Workload synthesizing.}
To the best of our knowledge, there is no open-sourced trace available for offline batch inference. 
Therefore, we synthesize our workloads by combining existing well-known single-modal traces, including two chatbot traces WildChat~\cite{zhao2024wildchat1mchatgptinteraction},
ShareGPT~\cite{sharegpt}, and two API services traces Azure-Trace~\cite{stojkovic2024dynamollmdesigningllminference}, BurstGPT~\cite{wang2024burstgpt},
one video generation trace OpenVid~\cite{nan2024openvid1mlargescalehighqualitydataset}
\footnote{We calculate the \outlen{} of a video generation request using the frames and quality of the videos in OpenVid.}, and one benchmark MMLU~\cite{hendrycks2021measuringmassivemultitasklanguage}.
Figure~\ref{fig:motivation-traces-length} illustrates the length distribution and \density{} of each trace. 
These single-modal traces have different representative characteristics: 
BurstGPT and Azure-Trace requests are highly compute-intensive, OpenVid requests are memory-intensive, while WildChat, ShareGPT have a mild \density{}. Besides, MMLU requests have high prefix sharing. 
We synthesize a variety of multi-modal workloads with different \psr{} and \density{} by combining different ratios of traces, based on which we demonstrate the effectiveness and generality of proposed \sys{}. Detailed methodology of synthetic workloads is described in Appendix~\refsec{sec:appendix:method-synthesize}.

\begin{table}[t!]
\centering
\small
\resizebox{\linewidth}{!}{
\begin{tabular}{|c|c|c|}
    \hline
    & High Prefix Sharing & Low Prefix Sharing \\ \hline
    Compute-intensive & Trace\#1 (1.4, 35\%) & Trace\#3 (1.4, 5\%) \\ \hline
    Memory-intensive & Trace\#2 (0.9, 35\%) & Trace\#4 (0.9, 5\%)\\ \hline
\end{tabular}
}
\caption{
Four representative synthesized workloads. Trace\#X (A,B\%) has a \density{} of A, with a \psr{} of B\%. For example, Trace\#1 is compute-intensive with high prefix sharing, which has a compute density of $1.4$ larger than $1$ and a \psr{} of $35$\%. Note that $35$\% is a high \psr{} as most workloads have less than $20$\% as shown in Table~\ref{tab:appendix-traces}.
Without losing genericity, Figure~\ref{fig:eval-sensitivity-workloads} shows more trace combinations and reports \sys{}'s performance on them. 
}
\vspace{-4mm}
\label{tab:eval-representative-workloads}
\end{table}

Table~\ref{tab:eval-representative-workloads} shows the four most representative workloads we mainly use in evaluation, which have different resource demands and \psr{}s.
Each synthesized workload is made from BurstGPT, MMLU, and OpenVid and contains at least $400,000$ requests, which require $5$ A100 GPU hours and are large enough to reach a stable performance. Evaluation results on more ratios are presented in \refsec{sec:eval:sensitivity:workloads}. We also present results with other combinations of traces in~\refsec{sec:appendix:extensive-evaluation}.

\begin{figure}[!t]
    \centering
    \includegraphics[width=\linewidth]{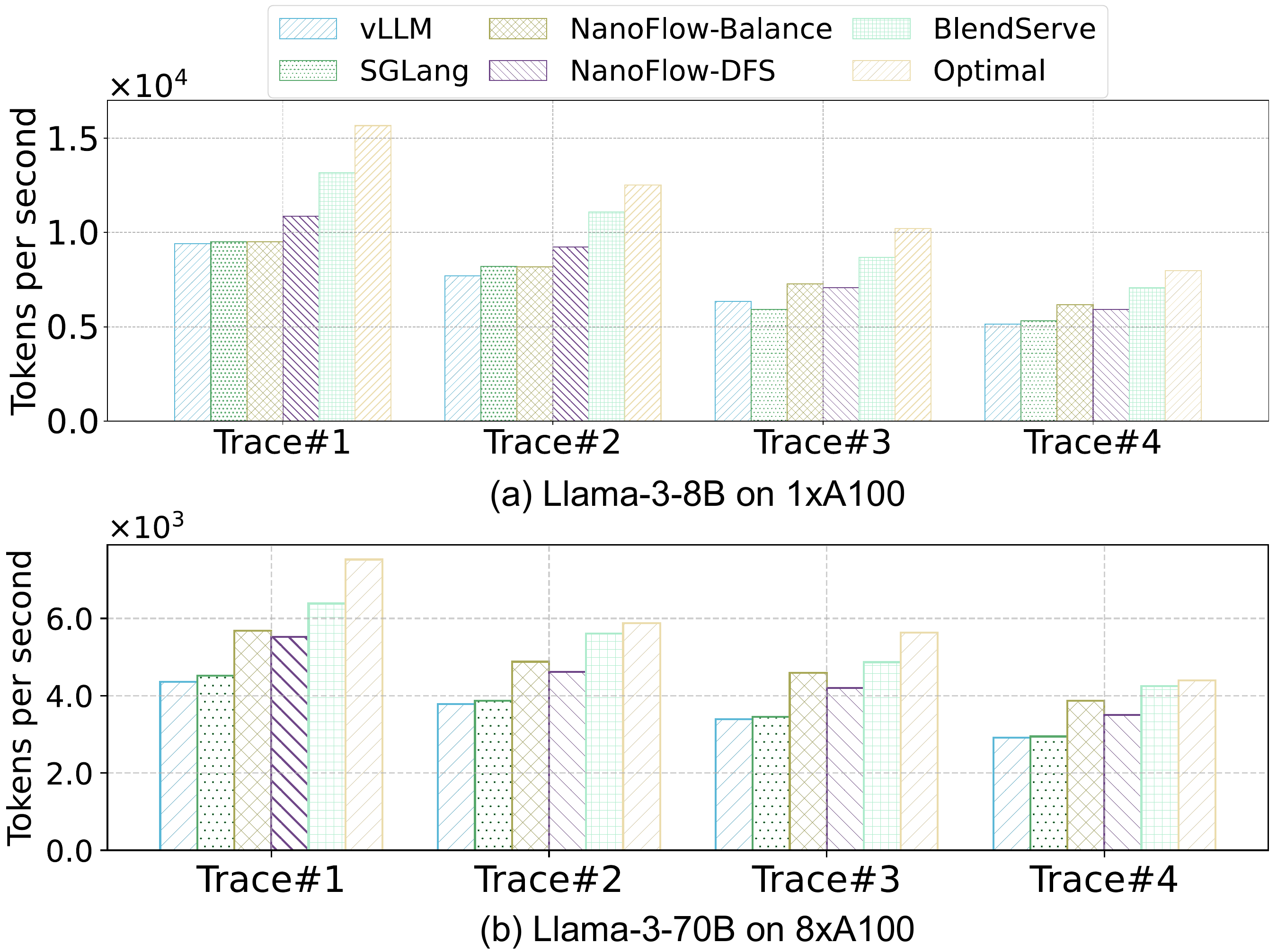}
    \caption{End-to-end throughput evaluation. \sys{} consistently outperforms baselines. \new{For \lm{}-3-8B, \sys{} achieves an average speedup of \avgimprovethansota{} compared to the best baseline, NanoFlow-DFS. For \lm{}-3-70B, \sys{} provides an average improvement of $18.6$\% over NanoFlow-DFS. Notably, \sys{} achieves \avgtoopt{} of optimal throughput on average.}}
    \label{fig:eval-e2e-tps}
    \vspace{-3mm}
\end{figure}

\MyPara{Models and hardware configurations.} 
\new{We evaluate \sys{} mainly with two widely-used open-sourced models, \lm{}-3.1-8B and \lm{}-3.1-70B~\cite{dubey2024llama3herdmodels}, on $1$ and $8$ A100 80GB SXM GPUs, respectively. 
To demonstrate the generality and robustness of \sys{}, we also evaluate models of different sizes with various numbers of GPUs, including Qwen-2.5-7B~\cite{bai2023qwentechnicalreport} and \lm{}-2-7B~\cite{touvron2023llama2openfoundation} on $1\times$A100, as well as Qwen-2.5-72B and DeepSeek-67B~\cite{deepseekai2024deepseekllmscalingopensource} on $8\times$A100. Due to the GPU resource limit, we conduct these experiments with a cycle-accurate simulator as discussed in~\refsec{sec:eval:sensitivity:workloads}.
For the distributed setting, we enable tensor parallelism with the degree of $8$ GPUs for all baselines.}

\MyPara{Baseline frameworks.} 
We use two widely used frameworks, vLLM~\cite{vllm} and SGLang~\cite{zheng2024sglangefficientexecutionstructured}, and a throughput-oriented framework, NanoFlow \cite{zhu2024nanoflowoptimallargelanguage}\footnote{We use vLLM v0.6.3.post2.dev102 (commit: e26d37a1) and SGLang v0.3.4.post1 (commit: 3f5ac88) as comparison baselines.}. 
\new{We also include a latency SLO-optimized framework, DistServe~\cite{zhong2024distservedisaggregatingprefilldecoding}, to compare P/D disaggregation in offline inference settings as detailed in~\refsec{sec-eval-subsec-e2e}}. 
We do not evaluate frameworks that are designed for resource-constrained settings, e.g., FlexGen~\cite{sheng2023flexgenhighthroughputgenerativeinference} and HeteGen~\cite{zhao2024hetegenheterogeneousparallelinference}.
For vLLM and SGLang, we enable prefix caching for both and reorder each workload trace into a DFS order, which can achieve a high \psr{}.
For NanoFlow, we add prefix caching support for fair comparison. 
For each workload trace, we evaluate the performance of NanoFlow using both DFS (NanoFlow-DFS) and random ordering (NanoFlow-Balance). The improvement of \sys{} over NanoFlow-DFS demonstrates the advantage of achieving resource balance, while the improvement over NanoFlow-Balance would highlight the benefit of a higher \psr{} as random ordering can achieve a relatively balanced resource. 
\new{Note that all baselines integrate \textit{continuous batching} which performs scheduling at request-level granularity, with the only difference being the ordering of requests.} 
As \sys{} focuses on improving GPU utilization, we do not measure CPU time to provide a fair comparison, including tokenizations, sampling, and scheduling~\cite{cpuoverhead}, for all baselines. We discuss the CPU overhead in \refsec{sec:appendix:cpu-overhead}.

\MyPara{Practical optimal throughput.}
To assess how closely \sys{}'s throughput approaches the optimal, we calculate optimal throughput with $T_o$ defined in \refsec{sec:formulate-optimal-throughput}. 
Due to the well-known performance interference issue in GPU hardware during spatial sharing~\cite{zhu2024nanoflowoptimallargelanguage, orion}, simply deriving $T_o$ with $\max(T_{comp},T_{mem})$ is impractical and unachievable. Therefore, to estimate a \textit{practical upperbound}, we employ a profiling-based approach similar to prior works~\cite{zhu2024nanoflowoptimallargelanguage,drift}.
\new{Specifically, instead of directly using $\max(T_{comp}, T_{mem})$ as the execution time, we profile the real execution time when overlapping GEMM with $T_{comp}$ and attention with $T_{mem}$, which is then used to calculate the practical upperbound of $T_o$.}

\subsection{End-to-end throughput}
\label{sec-eval-subsec-e2e}
\MyPara{Compared to existing frameworks.}
We measure the end-to-end throughput of \sys{} and all baselines, including vLLM-DFS, SGLang-DFS, NanoFlow-Balance, and NanoFlow-DFS.
We define end-to-end throughput as all processed tokens (including both input and output tokens) divided by the total processing time. 
For \lm{}-3-8B as shown in Figure~\ref{fig:eval-e2e-tps} (a), with a small \psr{} (i.e., Trace\#3 and \#4), NanoFlow-Balance works better than NanoFlow-DFS since resource overlapping contributes to more throughput gain. However, with a large \psr{}, NanoFlow-DFS achieves the highest throughput among all three baseline engines thanks to the high \psr{} and its operator-level resource overlapping. Since \sys{} is designed to leverage the best of both, it consistently outperforms the best baseline, NanoFlow-DFS, in all settings from $19.34$\% to $22.65$\%. Compared with vLLM-DFS, \sys{} achieves up to \upimprovethanvllm{} throughput speedup.
\new{For \lm{}-3-70B in Figure~\ref{fig:eval-e2e-tps} (b), \sys{} provides an average of $18.6$\% throughput improvement compared to NanoFlow-DFS, achieving $90.8$\% of practical optimal throughput. Note that NanoFlow provides higher throughput gain over vLLM compared to \lm{}-3-8B, due to the benefit of overlapping expensive communication operators with computation.}

\MyPara{Compared to practical optimal throughput.}
As shown in Figure~\ref{fig:eval-e2e-tps}, \sys{} achieves an average \avgtoopt{} and $90.8$\% of the optimal one on \lm{}-3-8B/70B, respectively. As there is a gap between the heuristic-based dual-scanner and the optimal scheduling, it is non-practical to achieve the optimal throughput which requires perfect resource overlapping on each step. Nevertheless, \sys{} still closes this gap to as low as 13\%, demonstrating its effectiveness in achieving both high \psr{} and high resource balance.

\begin{figure}[!t]
    \centering
    \includegraphics[width=\linewidth]{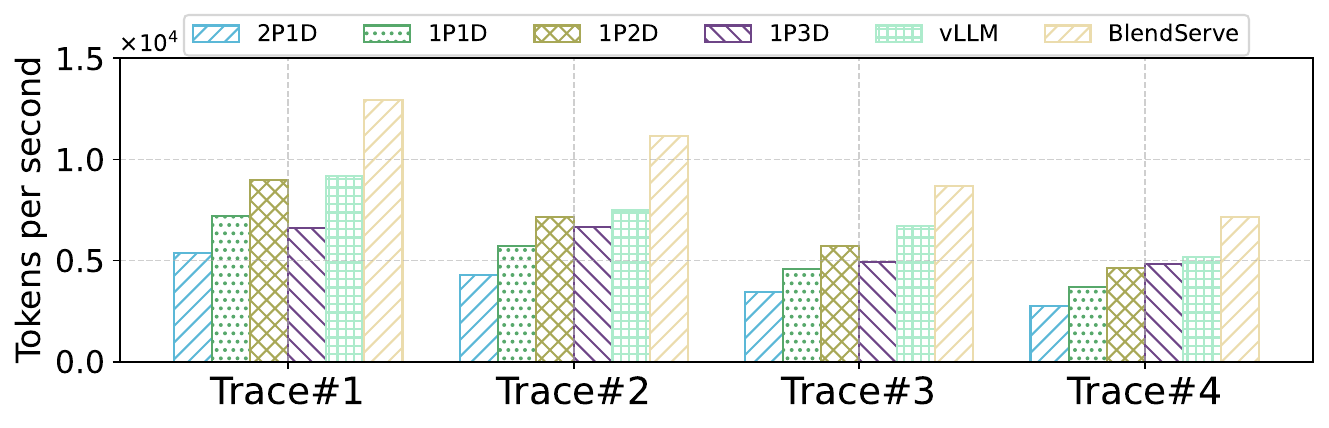}
    \caption{\new{End-to-end throughput (per GPU) evaluation when serving \lm{}-3-8B on $1\times$A100 GPU. \sys{} consistently outperforms baselines, including vLLM and P/D disaggregation. DistServe is less efficient when given more prefill clusters (e.g., 2P1D v.s. 1P2D) as selected workloads have more decode tokens.}}
    \label{fig:eval-distserve}
    \vspace{-2mm}
\end{figure}

\MyPara{Compared to P/D disaggregation.}
\new{We compare \sys{} with one popular design of P/D disaggregation, DistServe~\cite{zhong2024distservedisaggregatingprefilldecoding}, and cover several configurations including 1P1D, 1P2D, 2P1D, and 1P3D. Our implementation is based on SGLang where xPyD means x A100 GPUs are used as prefill clusters and y GPUs are used as decode clusters.
We collect the average per-GPU throughput when serving \lm{}-3-8B on A100 GPUs to provide a fair comparison, following the same workload and setup in~\refsec{sec:eval-setup}. As shown in Figure~\ref{fig:eval-distserve}, DistServe falls short on matching the throughput of vLLM under all configurations, which colocates prefill and decode. Despite being superior in latency-oriented settings where TTFT and TPOT could benefit from the disaggregated scaling and execution of prefill and decode, DistServe causes resource under-utilization due to the distinct resource usages of prefill and decode. Specifically, the memory bandwidth resources on prefill clusters are under-utilized by the compute-intensive prefill phases, and vice versa for compute resources in decode clusters.
}

\subsection{Performance analysis}
\label{sec:eval:performance-analysis}
We now ablate the key factors contributing to \sys{}'s performance improvement by showing \psr{} and hardware resource usage over time, corresponding to the two key design points introduced in \refsec{sec:formulate-optimal-throughput}. 

\begin{figure}[!t]
    \centering
    \includegraphics[width=0.95\linewidth]{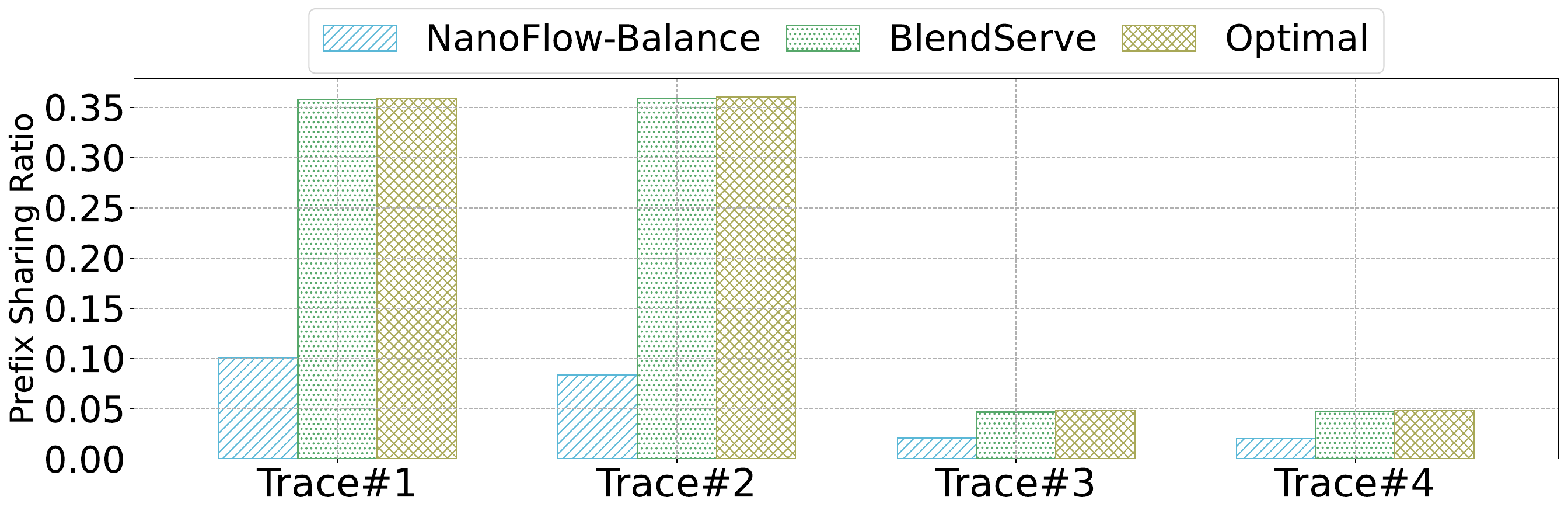}
    \caption{Prefix sharing ratio of four representative traces in the end-to-end evaluation. Note that the optimal value is measured via a DFS order of the prefix tree. \sys{} consistently maintains the benefit of prefix sharing, achieving $97$\% of maximal values.}
    \label{fig:eval-prefix}
    \vspace{-2mm}
\end{figure}

\begin{figure}[!t]
    \centering
    \includegraphics[width=\linewidth]{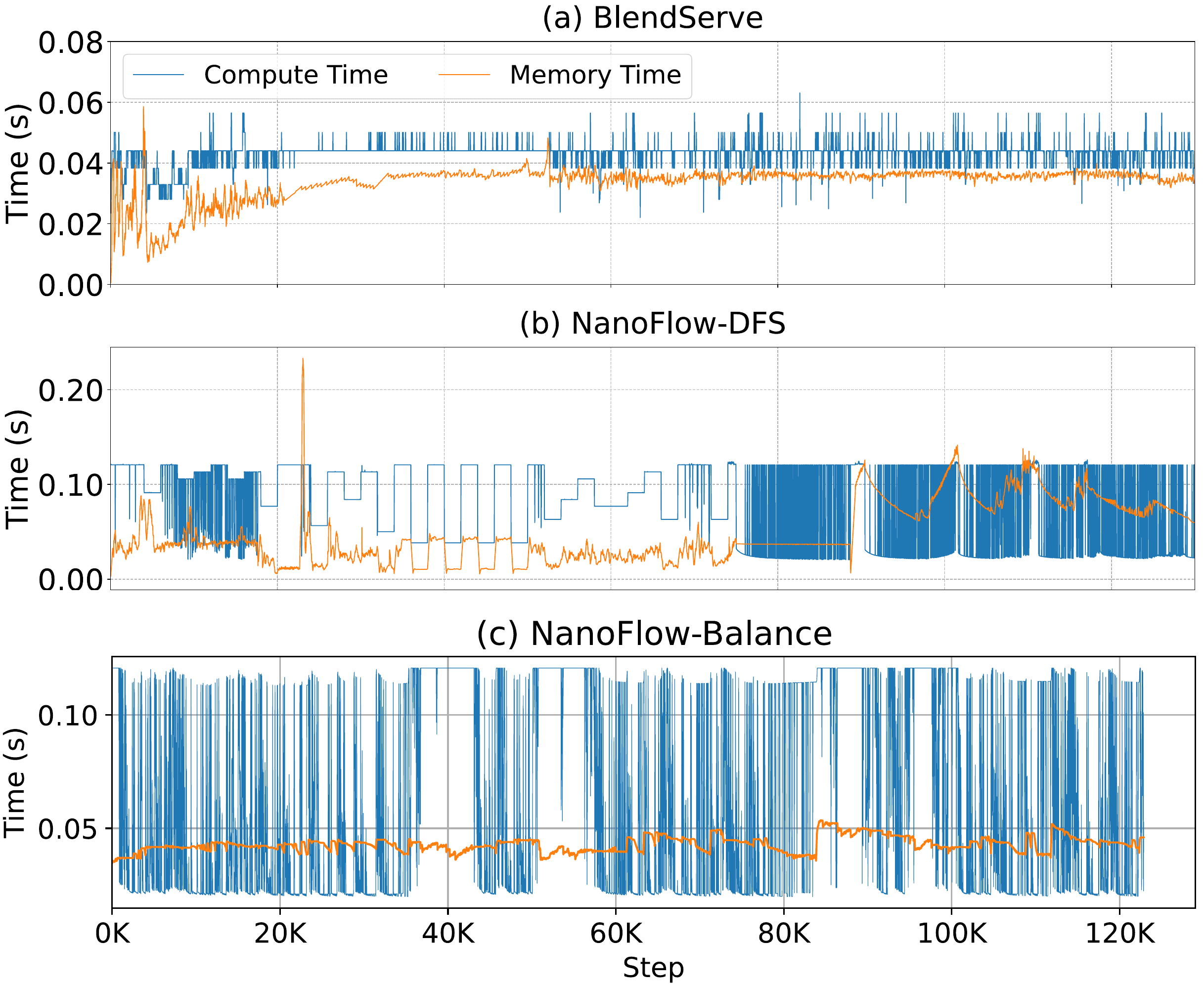}
    \caption{Compute and memory usages when serving Trace\#2. \sys{} well balances compute and memory time across steps and achieves consistently high resource utilization, whereas NanoFlow-DFS suffers from fluctuating compute and memory time and under-utilizes at least one type of resource at each step.}
    \label{fig:eval-resource}
\end{figure}

\MyPara{\Psr{}.}
To illustrate that \sys{} can achieve nearly optimal \psr{}, we collect the achieved \psr{} along with the maximal values. We manually exclude prefix sharing related to the recomputation of retracted requests. As shown in Figure~\ref{fig:eval-prefix}, \sys{} achieves over $97$\% of the optimal \psr{}. 
\new{In contrast, as the NanoFlow-Balance uses random ordering to interleave distinct requests without shared prefix locality, it fails below $30$\% of \psr{}. 
As a result, \sys{} provides an average of $1.36\times$ throughput improvement compared to NanoFlow-Balance with Trace\#1 and \#2.}

\MyPara{Hardware resource usage.}
To demonstrate how effectively \sys{} balances resource usage, we visualize the compute and memory usage of \sys{}, NanoFlow-DFS, and NanoFlow-Balance in Figure~\ref{fig:eval-resource}. We select Trace\#2, which has intensive memory usage and significant resource imbalance. 
\new{For each step, we collect the execution time of compute- and memory-bound operators. \sys{} maintains stable compute and memory usage, whereas NanoFlow-DFS exhibits significant fluctuations, resulting in resource under-utilization. For example, NanoFlow-DFS first under-utilizes memory bandwidth before $90K$ steps, then conducts excessive memory access. At the same time, NanoFlow-Balance achieves stable memory usage close to \sys{}. However, due to the massive recomputation and steep request length distribution, it still exhibits fluctuations in computation.}

\begin{figure}[!t]
    \centering
    \includegraphics[width=0.84\linewidth]{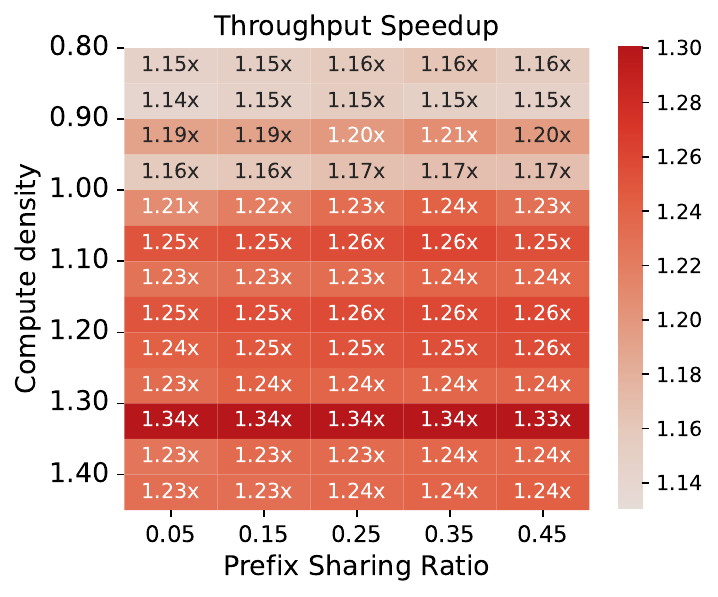}
    \caption{\textit{Simulated throughput} improvement of \sys{} compared to NanoFlow-DFS on workloads synthesized from BurstGPT, MMLU, and OpenVid. We use different numbers of requests from these traces to compose workloads with different \density{} and \psr{}. \sys{} consistently surpasses baselines, with an average of $1.23\times$ throughput improvement.}
    \vspace{-1mm}
    \label{fig:eval-sensitivity-workloads}
\end{figure}

\subsection{Sensitivity study}
\label{sec:eval:sensitivity:workloads}
To demonstrate the generality of \sys{} in real-world scenarios, we evaluate on more diverse synthetic workloads, with a large range of \density{} and \psr{}. In addition to the four most representative workloads shown in Table~\ref{tab:eval-representative-workloads}, we conduct a grid search of \density{} from $0.80$ to $1.40$ and \psr{} from $0.05$ to $0.45$ with step sizes $0.05$ and $0.10$, respectively. In total, we synthesize $65$ workloads to compare \sys{} and the best-performed baseline NanoFlow-DFS. Due to limited GPU resources, we use the frontend scheduler of \sys{} to generate actual batch schedules that are the same as running on real GPUs, which are then fed into a \textit{simulated GPU backend} to get the estimated inference time. For the backend simulation, we use polynomial fit to estimate the GPU runtime when given a certain amount of compute and memory usage. Our calibration shows only a $0.91$\% difference between the real and simulation speedup over the four representative workloads on average. Therefore, our simulation results practically reflect real performance. 

As shown in Figure~\ref{fig:eval-sensitivity-workloads}, \sys{} consistently outperforms the baseline in all workloads by $14$\% to $34$\%, with an average speedup of $22.53$\%. Since both \sys{} and NanoFlow-DFS achieve near-optimal \psr{}, the inference throughput remains stable when \psr{} varies. However, the benefits \sys{} gains from resource overlapping tend to shrink with smaller \density{}s, potentially due to more severe GPU interference on memory-intensive workloads. Additionally, the relative speedup achieves its maximum of $1.34\times$ when \density{} is around $1.30$, potentially because resource overlapping and GPU interference strike a balance under this ratio.

\subsection{Distributed deployment and other LLMs}
\label{sec:eval:distributed}
In this section, we evaluate \sys{}'s effectiveness and scalability in a distributed setting with data parallelism (DP).
In addition, we evaluate \sys{} on four other models, \new{including Qwen-2.5-7B, \lm{}-2-7B, Qwen-2.5-72B, and DeepSeek-67B, to show its general applicability}.

\begin{table}[t]
\centering
\small
\resizebox{\linewidth}{!}{
\begin{tabular}{|c|c|c|c|c|}
\hline
Tput & Trace\#1 & Trace\#2 & Trace\#3 & Trace\#4 \\
\hline
DP=1 & 11080 & 8408 & 8403 & 6325 \\
\hline
DP=2 & 20561 (1.85x) & 16261 (1.93x) & 15623 (1.85x) & 12246 (1.93x) \\
\hline
DP=4 & 41928 (3.78x) & 32537 (3.86x) & 32026 (3.81x) & 24541 (3.88x) \\
\hline
\end{tabular}
}
\caption{\new{\textit{Throughput scalability}} of \sys{} when serving \lm{}-3-8B with different DP sizes. \sys{} perfectly partitions requests among DP workers and scales near linearly.}
\label{tab:eval-dp}
\vspace{-4mm}
\end{table}

\begin{figure}[!t]
    \centering
    \includegraphics[width=\linewidth]{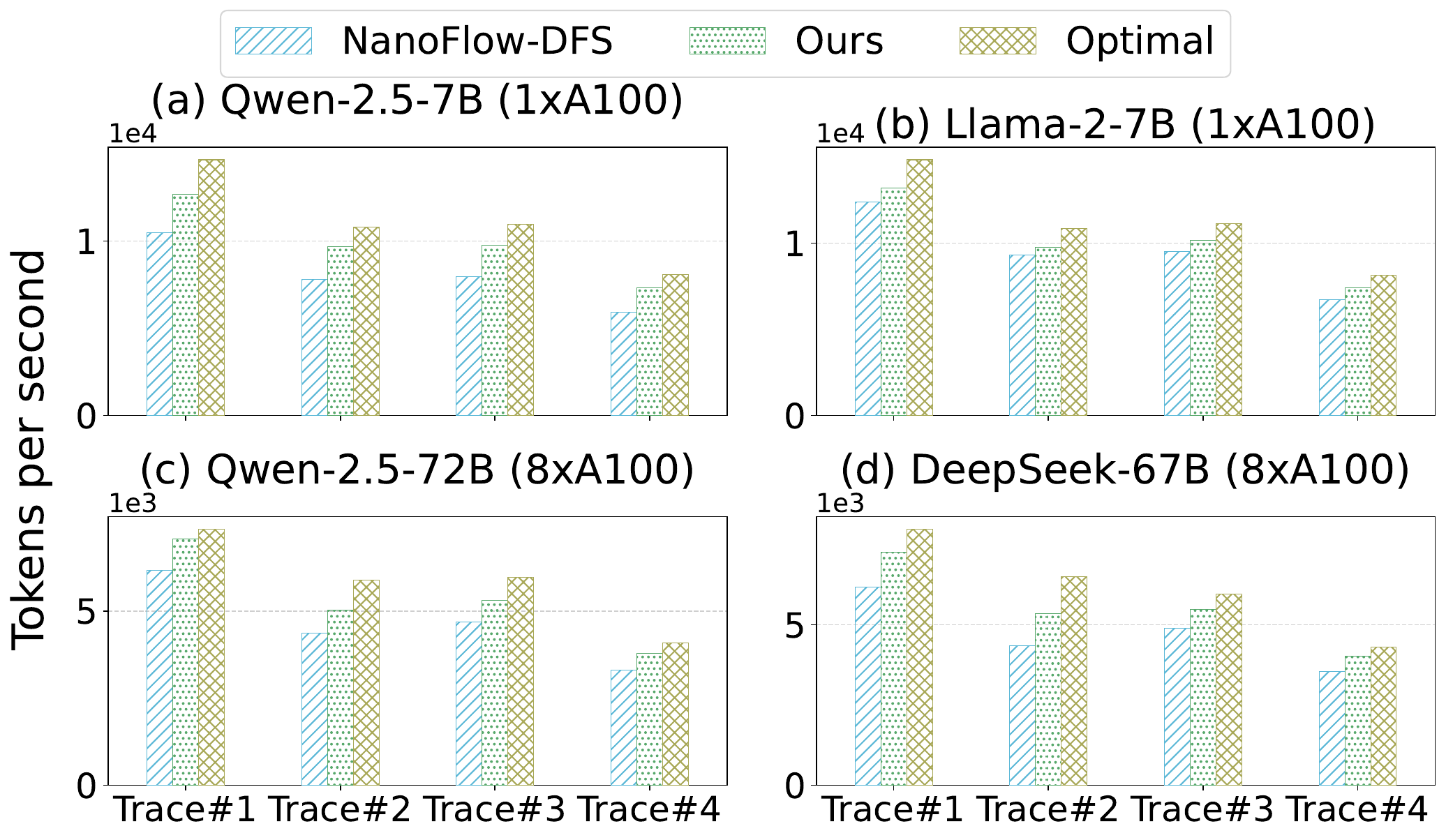}
    \caption{\textit{Simulated throughput} of \sys{} on different models with different number of GPUs. \new{\sys{} consistently surpasses the best baseline, NanoFlow-DFS, with up to $24.4$\% improvement over $4$ selected traces and models.}}
    \label{fig:eval-models}
\end{figure}

\MyPara{Data parallelism.}
We evaluate the strong scalability of \sys{} with various numbers of DP nodes by serving \lm{}-3-8B on A100 GPUs, following the design in~\refsec{sec:design:distribute} and the same workload setup in Table~\ref{tab:eval-representative-workloads}. 
As shown in Table~\ref{tab:eval-dp}, throughput increases linearly with the number of DP nodes.

\MyPara{Other LLMs.}
\new{We also evaluate \sys{} when serving Qwen-2.5-7B and \lm{}-2-7B on $1\times$ A100 GPU, as well as Qwen-2.5-72B and DeepSeek-67B on $8\times$ A100 GPUs as shown in Figure~\ref{fig:eval-models}. 
We redo the trace synthesis with the same recipe in~\refsec{sec:eval-setup}, as different models indicate different compute density.
Note that due to the GPU resource limit, we use the \textit{profile-guided simulation} as detailed in~\refsec{sec:eval:sensitivity:workloads} for this evaluation. 
Similarly, \sys{} improves throughput by an average of $15.2$\% compared to NanoFlow-DFS and achieves $89.9$\% of practical optimal throughput on average.}

\section{Discussion}
\label{sec:discussion}

\MyPara{Distributed parallelisms.}
\new{
\sys's design is generic to various parallelisms in distributed inference.
We have discussed data parallelism (DP) and tensor parallelism (TP) in~\refsec{sec:design:distribute}, and demonstrate its effectiveness in the evaluation.
In addition, \sys{} is compatible with various other parallelisms, including pipeline parallelism (PP), sequence parallelism (SP), and context parallelism (CP). 
For PP, as different pipeline stages will process identical batches sequentially while keeping each stage the same, \sys's scheduling can be directly adopted without modification. For SP~\cite{jacobs2023deepspeedulyssesoptimizationsenabling} and CP~\cite{liu2023ringattentionblockwisetransformers}, as attention and non-attention computation are sharded across SP/CP ranks, both the compute capability and memory bandwidth are scaled accordingly. Therefore, \sys{} is extended to SP/CP by including the scaled resources in the compute density calculation.}

\MyPara{Attention variants.}
\new{\sys{} is generic to attention variants, including MHA, MQA, GQA~\cite{ainslie2023gqatraininggeneralizedmultiquery}, and recently released MLA~\cite{deepseekai2024deepseekllmscalingopensource} and GLA~\cite{zadouri2025hardwareefficientattentionfastdecoding}, by considering the arithmetic intensity of the attention operator during compute density calculation. Specifically, \sys{} considers different variants by adapting the memory cost model $\texttt{Mem}(r)$ (\refsec{sec-analysis-request-density}) towards the real execution time. 
We have included \lm{}-2-7B with MHA, Qwen-2.5-7B with GQA (of group size $7$), and Llama-3-8B with GQA (of group size $4$) in our evaluation.}

\MyPara{End-to-end latency.}
\new{Given the same set of requests, \sys{} has the lowest worst turnaround latency across requests because it has the highest throughput compared to existing frameworks. Furthermore, \sys{} can ensure the latency requirement of offline batch inference by only blending requests within a specified time window. For example, \sys{} processes the previous X-hour request pool while queuing the next X-hour requests, moving to the subsequent X-hour window after completing the current one. 
}
\vspace{-4mm}

\section{Related Work}
\label{sec:related-work}

\MyPara{LLM serving optimizations.} 
Efficient LLM serving has been extensively studied for both online and offline scenarios. 
For online inference, Orca~\citep{orca}, vLLM~\citep{vllm}, SGLang~\citep{zheng2024sglangefficientexecutionstructured}, FastServe~\citep{wu2024fastdistributedinferenceserving}, and VTC~\citep{sheng2024fairnessservinglargelanguage} propose continuous batching, paged attention, prefix sharing, prefill-decode disaggregation, Multi-Level Feedback Queue scheduling, and Virtual Token Counter scheduling, respectively, to improve performance and/or fairness.
For offline inference, FlexGen~\citep{sheng2023flexgenhighthroughputgenerativeinference}, PowerInfer~\citep{powerinfer}, TwinPilots~\citep{twinpilots}, HeteGen~\citep{zhao2024hetegenheterogeneousparallelinference}, Fiddler~\citep{kamahori2024fiddlercpugpuorchestrationfast}, and NEO~\citep{jiang2024neosavinggpumemory} target \textit{resource-constrained} settings where GPU memory is insufficient. These methods extensively leverage CPUs to offload model weights, activations, KV-cache, and computation. However, due to limited GPU/CPU interconnect bandwidth, offloading introduces significant GPU underutilization, leading to low throughput. 
Unlike these approaches, \sys{} focuses on throughput-oriented offline inference with resource-aware batching. 

\MyPara{Resource overlapping techniques.}
Resource overlapping is a trendy approach to improve GPU utilization.
Rammer~\citep{rammer} introduces operator-level overlapping for deep neural network compilers. 
NanoFlow~\citep{zhu2024nanoflowoptimallargelanguage} extends operator-level overlapping to LLM serving. 
Sarathi-Serve~\cite{agrawal2024tamingthroughputlatencytradeoffllm} and FastGen~\citep{holmes2024deepspeedfastgenhighthroughputtextgeneration} apply phase-level overlapping to LLM serving. 
MuxServe~\citep{duan2024muxserveflexiblespatialtemporalmultiplexing} colocates models based on their popularity and resource characteristics, targeting resource-limited scenarios. 
Compared to them, \sys{} is the first to exploit request-level resource overlapping with request reordering.

\section{Conclusion}
\label{sec:conclusion}

We present \sys{}, an \offinf{} system that maximizes both compute-memory overlapping and prefix sharing for near-optimal throughput. \sys{} exploits the relaxed latency objective in offline batch inference and reorders compute- and memory-intensive requests through a resource-aware prefix tree and a dual scanner searching algorithm. 
\sys{} achieves up to \upimprovethanvllm{} higher throughput over vLLM and SGLang and 90\% of the optimal throughput.

\appendix
\section{Appendix}
\label{sec:appendix}

\subsection{Pseudoscope for node sort, split, and dual scan}
\label{sec:appendix-pseudo-algorithm}

\begin{algorithm}[ht]
\footnotesize
\caption{Layer-wise Sorting}
\label{alg:design-transformation}
\begin{algorithmic}[1]

\Function{layer\_sort}{$ptr$}    %
    \If{$ptr$ is not leaf node}
        \State \textbf{sort} $ptr.childList$ based on \density
        \For{$cptr \in ptr.childList$}
            \State \Call{layer\_sort}{$cptr$}
        \EndFor
    \EndIf
\EndFunction

\end{algorithmic}
\end{algorithm}

\begin{algorithm}[ht]
\footnotesize
\caption{Node Splitting}
\label{alg:design-split}
\begin{algorithmic}[1]

\State \textbf{Initialize} $leaf\_list \gets \{\}$

\Function{node\_split}{$ptr, t$}
    \State $ptr.len_{prefix} \gets \text{length of prefix to } ptr$
    \If{$ptr.len_{prefix} \cdot \text{len}(ptr.childList) > t$}
        \State $ptr.len_{prefix} \gets ptr.len_{prefix} - ptr.len$
        \State \Call{update\_subtree\_density}{$ptr$}
        \State \textbf{append} $ptr$ \textbf{to} $leaf\_list$
    \Else
        \For{$cptr \in ptr.childList$}
            \State \Call{node\_split}{$cptr,\;\frac{t}{\text{len}(ptr.childList)}$}
        \EndFor
    \EndIf
    \If{$ptr$ is root node}
        \State \textbf{sort} $leaf\_list$ based on \density
    \EndIf
\EndFunction

\end{algorithmic}
\end{algorithm}

\begin{algorithm}[ht]
\footnotesize
\caption{\new{Dual Scan}}
\label{alg:dual-scan}
\begin{algorithmic}[1]
\Function{\new{dual\_scan}}{\new{$\rho(rt),\;\rho(L),\;\rho(R),\;M$}}
    \State \textbf{Input:} compute density of root $\rho(rt)$, left child $\rho(L)$, and right child $\rho(R)$; total available GPU memory $M$
    \State \textbf{Output:} chunked prefill budgets for the left child $C_L$, and right child $C_R$ (in terms of tokens)
    
    \Statex \# Step 1: partition memory $M$ according to the compute density
    \State $M_L \gets M \cdot \dfrac{\rho(rt)-\rho(R)}{\rho(L)-\rho(R)}$
    \State $M_R \gets M \cdot \dfrac{\rho(L)-\rho(rt)}{\rho(L)-\rho(R)}$

    \Statex \# Step 2: calculate the chunked prefill budget according to the
    \Statex \# \qquad \quad estimated \inlen{} $p_L$ and \outlen{} $d_L$
    \State $N_L \gets \dfrac{M_L}{(p_L + d_L/2)\cdot H_{kv}\cdot L\cdot4}$ \# number of decode requests
    \State $C_L \gets N_L \cdot \dfrac{p_L}{d_L}$ \# scale into prefill token budget

    \Statex \# Step 3: calculate the chunked prefill budget of the right child
    \State $N_R \gets \dfrac{M_R}{(p_R + d_R/2)\cdot H_{kv}\cdot L\cdot4}$ 
    \State $C_R \gets N_R \cdot \dfrac{p_R}{d_R}$

    \State \Return $(C_L,\, C_R)$ \# determines number of requests that are admitted
\EndFunction

\end{algorithmic}
\end{algorithm}

\subsection{Implementation details}
\label{sec:appendix:impl-details}

We introduce additional noteworthy details of our implementation in \sys{} here.

\MyPara{Offline prefix tree.} We preprocess all requests and construct a prefix tree following a Trie Tree to capture their shared prefixes before serving. 
After compute density calculation and node sorting, we merge sub-trees into single nodes if doing so does not hurt the \psr{}. This merging reduces fragmentation which would cause fluctuation during the dual scanner process.

\MyPara{Runtime prefix tree.} The runtime prefix tree in \sys{} is implemented based on SGLang~\cite{zheng2024sglangefficientexecutionstructured}. It manages runtime information related to prefix sharing, including a dynamic Trie Tree and a mapping between the physical memory and key-value tokens. We also employ intra-batch prefix sharing, enabling exactly-once computation of shared prefixes for a single batch, which is particularly beneficial for offline processing using a DFS order.

\MyPara{Batch scheduler.} The batch scheduler within the dual scanner is implemented following NanoFlow~\cite{zhu2024nanoflowoptimallargelanguage}. It strictly enforces batch sizes in multiples of 128 to ensure higher hardware utilization. We also incorporate chunked prefill and continuous batching following state-of-the-art serving systems~\cite{agrawal2024tamingthroughputlatencytradeoffllm,orca}. 

\MyPara{Backend engine.} Our backend engine is built in C++ following NanoFlow’s operator-level overlapping approach, which enables simultaneous execution of compute-intensive operators like GEMM and memory-intensive operators like self-attention~\cite{zhu2024nanoflowoptimallargelanguage}. Based on the operator-level overlapping, \sys{} overlaps operators from requests with distinct resource usages.

\subsection{Methodolody of workload synthesize}
\label{sec:appendix:method-synthesize}
To synthesize workloads that reflect real use cases, we collect a variety of open-source inference traces that have distinct characterization, including \density{}, \psr{}, and modalities. We illustrate their length distribution in Figure~\ref{fig:motivation-traces-length}. For each set of traces, we add a unique system prompt ahead of prompts as it is not collected. For traces without detailed prompt content, we randomize their prompts' token ids corresponding to their prompt length. For video generation requests, we use OpenVid~\cite{nan2024openvid1mlargescalehighqualitydataset} and treat the videos in training datasets as their auto-regressive generation output. For each video, we collect its \outlen{} by counting the number of frames and multiplying it by $256$, which represents the number of tokens per frame observed in normal videos~\cite{xue2024longvilascalinglongcontextvisual, liu2024worldmodelmillionlengthvideo}. 
Additionally, we normalize the average \outlen{} of OpenVid to $16$K as the original $45$K is too large for evaluation of \lm{}-3.1-8B on a single A100 GPU. We also normalize the average \outlen{} of WildChat~\cite{zhao2024wildchat1mchatgptinteraction} to $256$ for a more compute-intensive workload while maintaining the length variance. We calculate the resource characterization in Table~\ref{tab:appendix-traces}.

\begin{table}[h!]
\centering
\small
\resizebox{\linewidth}{!}{
\begin{tabular}{l|c|c|c|c|c|c}
\toprule
 & ShareGPT & WildChat & Azure-Trace & OpenVid & BurstGPT & MMLU \\
\midrule
Prefix sharing & 0.02 & 0.19 & 0.01 & 0.00 & 0.02 & 0.86 \\
Compute density      & 3.12 & 2.13 & 33.2 & 0.05 & 17.78 & 54.91 \\
\bottomrule
\end{tabular}
}
\caption{\Psr{} and \density{} of collected traces. OpenVid is memory-intensive due to its large \outlen{}, while MMLU has a high \psr{} of $86.46$\%. Others are compute-intensive with less \psr{}.}
\label{tab:appendix-traces}
\end{table}

To cover the real cases in \offinf{}, we conduct a grid search of synthetic workloads with different \density{} and \psr{}. To reach the desired \density{} $t$, we combine one compute-intensive trace among ShareGPT, Azure-Trace, WildChat, and BurstGPT, and a memory-intensive video generation trace OpenVid. Based on $t$ and \density{} of selected traces, we calculate the required request number of each trace, with a total number of $40,000$ requests. Then we mix requests from MMLU to reach the desired number of \psr{} to get the synthetic workload. Such a synthetic workload has a diverse request length and various resource characterization, which is similar to real-world cases.

\subsection{Extensive evaluation of synthetic workloads}
\label{sec:appendix:extensive-evaluation}
\begin{figure}[!t]
    \centering
    \includegraphics[width=0.80\linewidth]{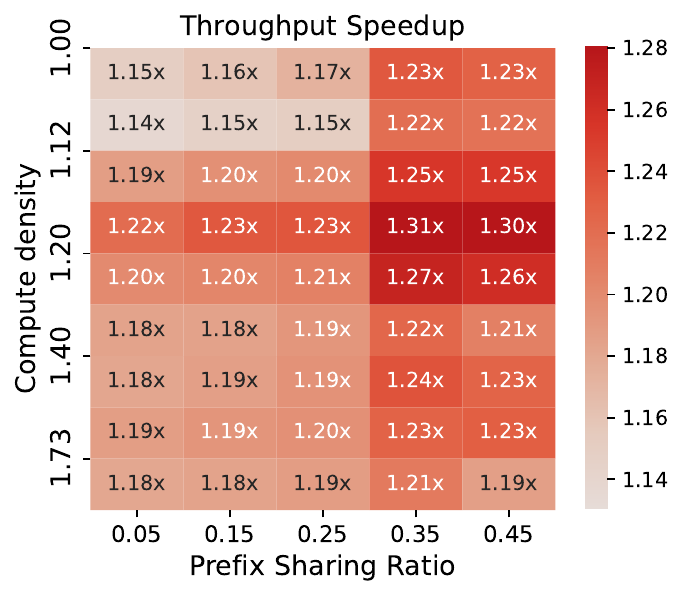}
    \caption{\textit{Simulated throughput} improvement of \sys{} compared to NanoFlow-DFS on workloads synthesized from Azure-Trace, MMLU, and OpenVid. \sys{} achieves up to $31$\% throughput gain compared to baselines.}
    \label{fig:appendix-azure}
\end{figure}

\begin{figure}[!t]
    \centering
    \includegraphics[width=0.80\linewidth]{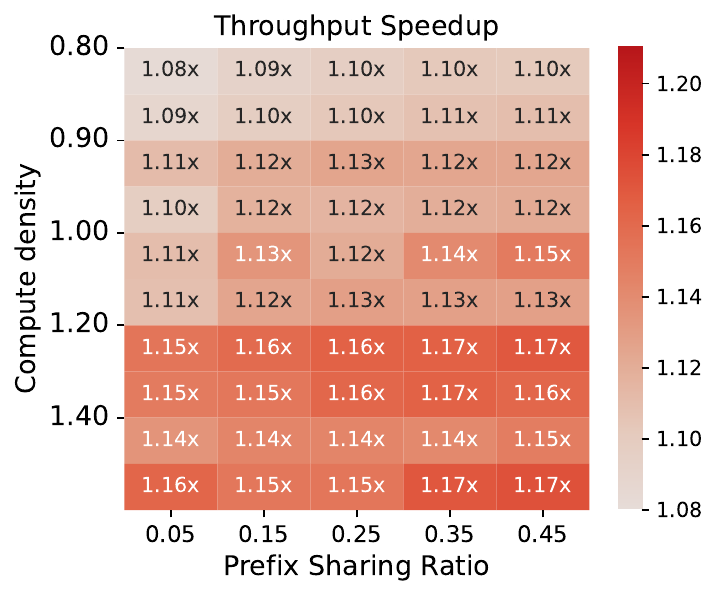}
    \caption{\textit{Simulated throughput} improvement of \sys{} compared to NanoFlow-DFS on workloads synthesized from ShareGPT, MMLU, and OpenVid. \sys{} consistently surpasses baselines by up to $17$\% throughput.}
    \label{fig:appendix-sharegpt}
\end{figure}

\begin{figure}[!t]
    \centering
    \includegraphics[width=0.80\linewidth]{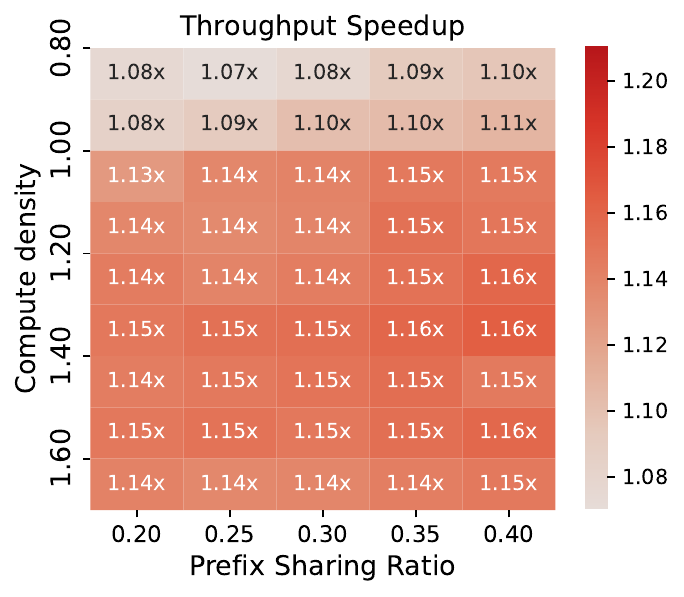}
    \caption{\textit{Simulated throughput} improvement of \sys{} compared to NanoFlow-DFS on workloads synthesized from WildChat, MMLU, and OpenVid.}
    \label{fig:appendix-wildchat}
\end{figure}

In addition to the main evaluations conducted on BurstGPT, MMLU, and OpenVid in~\refsec{sec:eval}, we also evaluate \sys{} on Azure-Trace (Figure~\ref{fig:appendix-azure}), ShareGPT (Figure~\ref{fig:appendix-sharegpt}), and WildChat (Figure~\ref{fig:appendix-wildchat}) to demonstrate the generality of proposed methods over diverse workloads, following the same experiment setup (\refsec{sec:eval-setup}).

Results show that \sys{} consistently surpasses baselines by $1.08\times$ to $1.31\times$ in different workloads. We find that \sys{} works better on BurstGPT and Azure-Trace due to their smaller variance of \outlen{}. When the \outlen{} variance is large in ShareGPT and WildChat, the sampling strategy works less effectively, leading to sub-optimal performance. We leave the better strategy for workloads with large variance \outlen{} that cannot be effectively captured by the prefix tree for future work.

\subsection{Scheduling overhead of \sys{}}
\label{sec:appendix:cpu-overhead}

As described in \refsec{sec-design}, \sys{} has two scheduling overhead: 
1) preprocessing all token ids of requests prompt to construct the prefix tree, followed by a series of tree transformations; and 2) runtime scheduling request batches based on the double scanner algorithm and the prefix tree to manage \kv{} memory. We now demonstrate that these two parts have minimal overhead compared to the GPU time.

\MyPara{Preprocessing overhead.}
There is no additional overhead for tokenization, since it is also necessary for model inference, and the storage for generated token ids is at the same magnitude as the input strings. Assuming $N$ requests with $T$ tokens in the prompts, for the trie tree construction with $D$ max depth, the time complexity $O(N\times D)$. Since requests' prompts diverge quickly, $D$ is typically small. In our evaluations, this process typically takes several minutes, which is negligible compared to hours of GPU inference.

\MyPara{Runtime scheduling overhead.}
Since the runtime batch size is typically at the magnitude of thousands, the runtime prefix tree is much smaller compared to the offline prefix tree built during preprocessing. Based on our measurement in evaluations, the operations on the runtime prefix tree take $0.08$~ms on average, with a P99 latency of $0.23$~ms, which is generally less than $10$\% compared to the GPU time. Such small runtime scheduling overhead can be effectively overlapped with asynchronous CPU scheduling, incurring zero overhead in end-to-end performance~\cite{zhu2024nanoflowoptimallargelanguage}.

\section{Artifact}

\subsection{Abstract}
This artifact provides an implementation of the proposed system using pre-built Docker images that encapsulate the codebase and runtime environment. 
All experiments are orchestrated through a single entry-point script for ease of use and automation. Experimental results are collected and visualized using a Jupyter notebook.

\subsection{Artifact check-list (meta-information)}
{\small
\begin{itemize}
  \item {\bf Algorithm: Offline inference schedule}
  \item {\bf Program: Python, C++}
  \item {\bf Compilation: nvcc, g++}
  \item {\bf Model: Meta-Llama-3-8B}
  \item {\bf Data set: Huggingface datasets}
  \item {\bf Hardware: A100-SXM4-80GB}
  \item {\bf Metrics: Tokens per second, prefix hit rate}
  \item {\bf How much disk space required (approximately)?: 50GB}
  \item {\bf How much time is needed to prepare workflow (approximately)?: 10mins}
  \item {\bf How much time is needed to complete experiments (approximately)?: 50 A100 hours}
  \item {\bf Publicly available?: Yes}
  \item {\bf Code licenses?: Apache-2.0 license}
  \item {\bf Data licenses?: Apache-2.0 license}
\end{itemize}
}

\subsection{Description}

\subsubsection{How to access}

A Docker image, including all software dependencies (compiled), model weights, code references, and scripts, is provided via a public \href{https://drive.google.com/file/d/1fVe8C1_tSAjiTW9UZ0Fsz7udHuCHiy2U/view?usp=sharing}{Google Drive link}. 
We also provide an image (without CUDA dependency) for reproducing subsets of experiments without the GPU backend via this \href{https://drive.google.com/file/d/1WG5H6JMOwBfdf47TGLNdY3gC3iWhA-r-/view?usp=sharing}{Google Drive link}.

\subsubsection{Hardware dependencies}
All evaluations are conducted with NVIDIA A100-SXM4-80GB GPUs.

\subsubsection{Software dependencies}
The desired environmental setup follows the official Docker container, i.e., \href{https://catalog.ngc.nvidia.com/orgs/nvidia/containers/nvhpc?version=23.11-devel-cuda_multi-ubuntu22.04}{23.11-devel-cuda\_multi}. 
The software libraries, including vLLM and NanoFlow, are also provided along with the image.

\subsubsection{Data sets}
The evaluated workloads are synthesized by combining several open-sourced traces with distinct characteristics, including \href{https://huggingface.co/datasets/nkp37/OpenVid-1M}{OpenVid-1M}, \href{https://github.com/HPMLL/BurstGPT?tab=readme-ov-file}{BurstGPT}, and \href{https://arxiv.org/abs/2009.03300}{MMLU}.

\subsubsection{Models}
Both Qwen-2.5-7B and LLama-3-8B are evaluated on A100 with TP=1, while Qwen-2.5-72B and Llama-3-70B are evaluated with TP=8. 
We mainly provide automated scripts for reproducing 8B models due to resource constraints, while others can be done in a similar way. 
\subsection{Installation}

We provide a pre-built Docker image that encapsulates all required dependencies.
Users should first download the image archive and load it into the local Docker
environment, then launch a container with the provided configuration.

\begin{Verbatim}[breaklines,fontsize=\small]
docker load -i blendserve.tar
docker run -it --gpus all \
  --name blendserve \
  -v /dev/shm:/dev/shm \
  blendserve:latest
\end{Verbatim}

After launching the container, the working directory is set to
\texttt{/root/blendserve}, which contains the full source code and scripts needed
to reproduce our results.

Some datasets and model weights are hosted on Hugging Face and require user
authentication. Please log in using the Hugging Face CLI with a valid access
token:

\begin{Verbatim}[breaklines,fontsize=\small]
hf auth login --token $YOUR_TOKEN
\end{Verbatim}

Detailed guidelines are provided in
\texttt{./README.md}. The main entry point for running experiments
is the script located at \texttt{./scripts/run.sh}.

\subsection{Evaluation and Expected Results}

All experiments are orchestrated through a single entry-point located at \texttt{./scripts/run.sh}, which sequentially launches the full set of experiments used in our evaluation. For convenience and flexibility, each experiment can also be executed independently by invoking the corresponding commands in the script. For each experiment, the raw outputs and aggregated results are stored in the corresponding experiment directory. Quantitative results are summarized in \texttt{combine.csv}, while visualizations and plots are generated using the provided Jupyter notebook \texttt{plot.ipynb}.

\bibliographystyle{ACM-Reference-Format}
\bibliography{_main.bib}

\end{document}